\documentclass[10pt,twocolumn,letterpaper]{article}
\usepackage[pagenumbers]{cvpr} % To force page numbers, e.g. for an arXiv version
\usepackage{multicol}
\usepackage{multirow}
\usepackage{graphicx}
\usepackage{bbm}
\usepackage{bm}
\usepackage{array}
\usepackage{makecell}
\usepackage{cuted}
\usepackage{capt-of}

\definecolor{cvprblue}{rgb}{0.21,0.49,0.74}
\usepackage[pagebackref,breaklinks,colorlinks,allcolors=cvprblue]{hyperref}

\title{C-DiffSET: Leveraging Latent Diffusion for SAR-to-EO Image Translation\\with Confidence-Guided Reliable Object Generation}

\author{Jeonghyeok Do\\
[0.3em]
KAIST\\
{\tt\small ehwjdgur0913@kaist.ac.kr}
\and
Jaehyup Lee \footnotemark[2]\\
[0.3em]
KNU\\
{\tt\small jaehyuplee@knu.ac.kr}
\and
Munchurl Kim \footnotemark[2]\\
[0.3em]
KAIST\\
{\tt\small mkimee@kaist.ac.kr}
\and
\small{\url{https://kaist-viclab.github.io/C-DiffSET_site}}
}

\begin{document}
\maketitle

\begin{strip}\centering
    \vspace{-1.5cm}
    \includegraphics[width=1.0\linewidth]{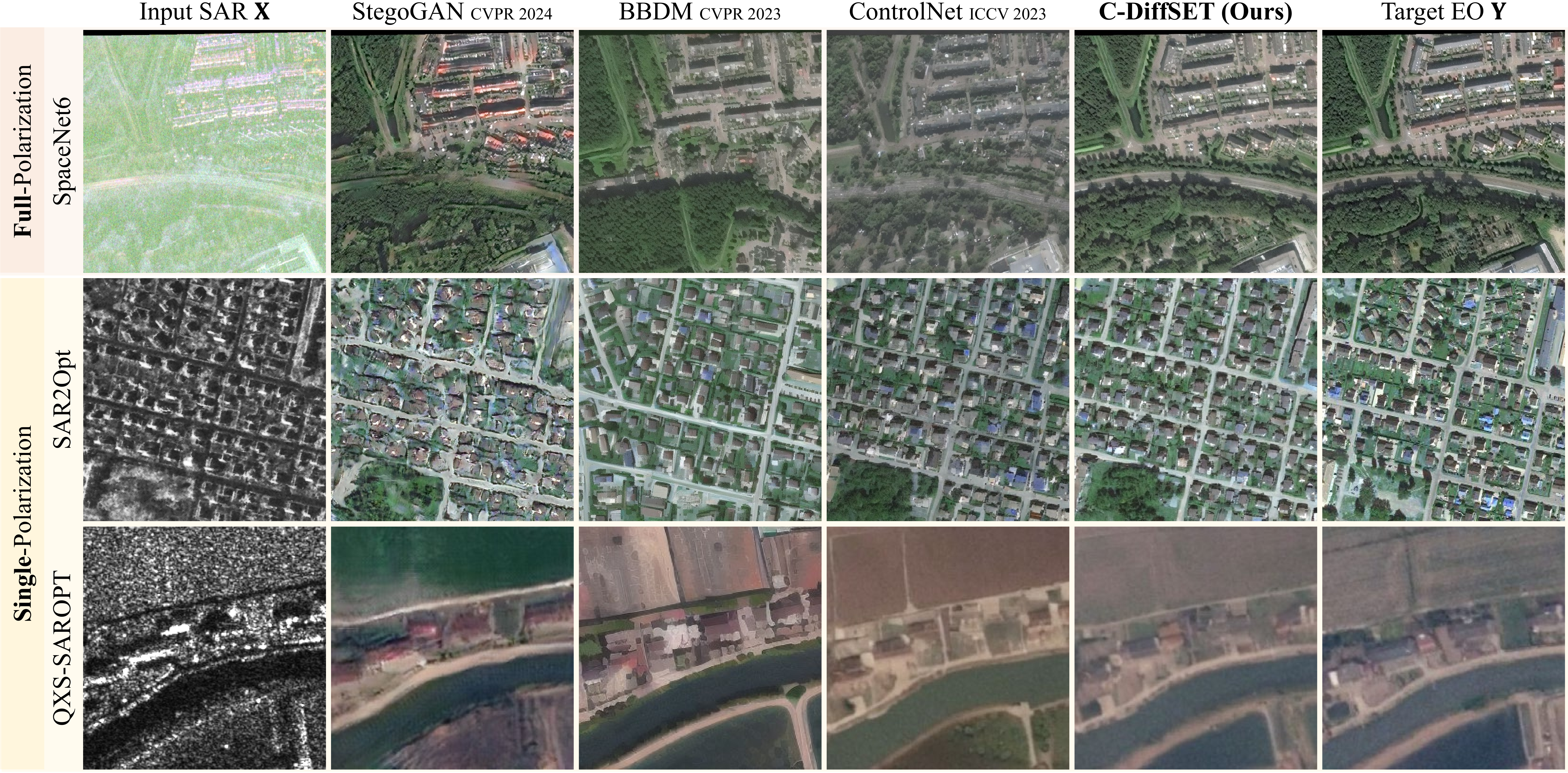}
    \vspace{-0.6cm}
    \captionof{figure}{Qualitative comparison of SAR-to-EO image translation (SET) results for very recent methods and our proposed C-DiffSET. The first row shows the SET results for full-polarization (HH, HV, VH, and VV) SAR input (ground sample distance (GSD) = 0.5$m$) of SpaceNet6 dataset, while the second and third rows exhibit the SET results for single-polarization (HH or VV) SAR input (GSD = 1$m$) of SAR2Opt and QXS-SAROPT datasets, respectively. As shown, our proposed C-DiffSET achieves superior structural accuracy and visual fidelity compared to very recent methods. Note that (i) the SET results for SpaceNet6 show better visual qualities than those of SAR2Opt and QXS-SAROPT because of using the full-polarization SAR input of SpaceNet6, and (ii) the SET results for SAR2Opt are of relatively higher visual qualities than those for QXS-SAROPT due to their less noisy SAR samples of SAR2Opt.}
    \label{fig:first}
\end{strip}

{
  \renewcommand{\thefootnote}%
    {\fnsymbol{footnote}}
  \footnotetext[2]{Co-corresponding authors (equal advising).}
}
\vspace{-5mm}

\begin{abstract}
Synthetic Aperture Radar (SAR) imagery provides robust environmental and temporal coverage (e.g., during clouds, seasons, day-night cycles), yet its noise and unique structural patterns pose interpretation challenges, especially for non-experts. SAR-to-EO (Electro-Optical) image translation (SET) has emerged to make SAR images more perceptually interpretable. However, traditional approaches trained from scratch on limited SAR-EO datasets are prone to overfitting. To address these challenges, we introduce Confidence Diffusion for SAR-to-EO Translation, called C-DiffSET, a framework leveraging pretrained Latent Diffusion Model (LDM) extensively trained on natural images, thus enabling effective adaptation to the EO domain. Remarkably, we find that the pretrained VAE encoder aligns SAR and EO images in the same latent space, even with varying noise levels in SAR inputs. To further improve pixel-wise fidelity for SET, we propose a confidence-guided diffusion (C-Diff) loss that mitigates artifacts from temporal discrepancies, such as appearing or disappearing objects, thereby enhancing structural accuracy. C-DiffSET achieves state-of-the-art (SOTA) results on multiple datasets, significantly outperforming the very recent image-to-image translation methods and SET methods with large margins.
\end{abstract}    
\section{Introduction}
\label{sec:intro}
Satellite imagery plays a critical role in various applications, including surveillance, transportation, agriculture, disaster assessment, and environmental monitoring \cite{sar_sur1, sar_trans1, sar_disa1, sar_moni1, sar_moni2, sebasti, sar_cloud1, sar_cloud2}. A significant portion of these applications relies on Electro-Optical (EO) imagery, which provides multi-spectral data captured by EO sensors. However, a fundamental limitation of EO imagery is its susceptibility to weather and lighting conditions, reducing its usability in scenarios such as cloudy weather or nighttime. In contrast, Synthetic Aperture Radar (SAR) offers robust sensing capabilities under all weather conditions without the need for light, making it ideal for nighttime operations. Despite these advantages, SAR imagery is generally very difficult to directly interpret the structural and contextual information due to the very different nature of its image structures with high speckle noises \cite{goodman, speckle_noise, character_zhang, character_soergel}, unlike the natural color images. Therefore, for easy interpretation and downstream tasks such as target detection and recognition \cite{lv1, lv2, lv4, object_detection_1, object_detection_2} developed for natural color images or satellite EO-RGB images, the SAR-to-EO image translation (SET) \cite{zhang, QXS, MCGAN, CFCASET, SGCLSET, DSE, CMdiffusion, cBBDM} has been demanded for the usability expansion for SAR image applications.  In SET, most existing approaches, including ours, focus on generating RGB bands of EO images \cite{zhang, QXS, MCGAN, DSE, CMdiffusion, cBBDM}, as RGB representations are widely used for both human perception and vision-based deep learning models. Therefore, the SET problem in this paper is focused on generating the RGB bands of EO images for SAR input images.

\begin{figure}[tbp]
  \centering
  \includegraphics[width=1.0\columnwidth]{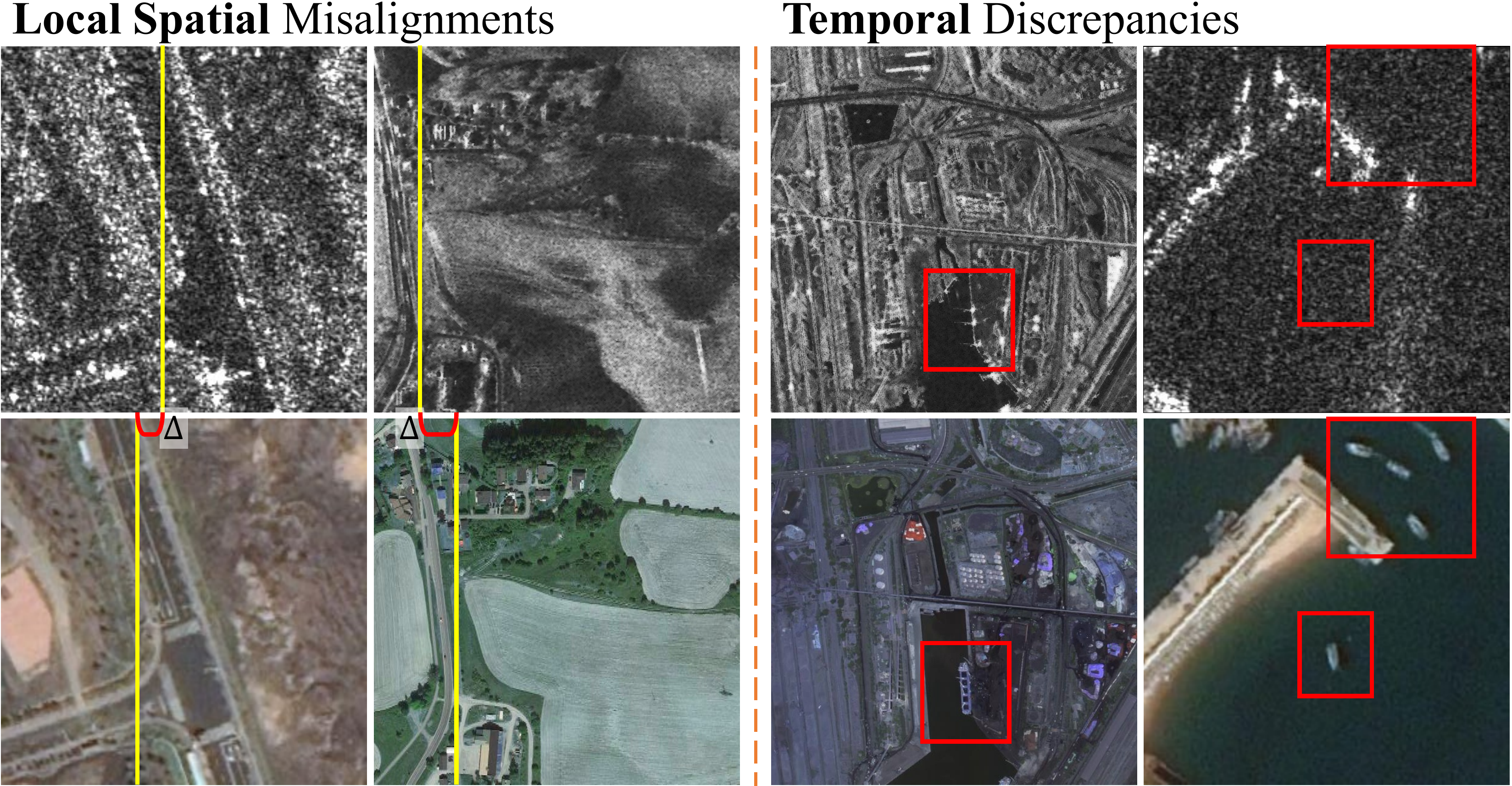}
  \vspace{-0.6cm}
  \caption{Examples of misalignments and discrepancies in paired SAR-EO datasets. \textit{Left}: Local spatial misalignments caused by sensor differences or acquisition conditions. \textit{Right}: Temporal discrepancies where objects (e.g., ships) appear or disappear between SAR and EO images due to their different acquisition times.}
  \label{fig:challenge}
  \vspace{-0.5cm}
\end{figure}

Recent advancements \cite{zhang, QXS, MCGAN, CFCASET, SGCLSET, DSE, CMdiffusion, cBBDM} have demonstrated the potential of generating EO-like outputs from SAR images, offering improved textures and enhanced color information. However, several challenges persist: (i) The domain gap between SAR and EO images makes SET an ill-posed problem; (ii) The misalignment between SAR and EO datasets is prone to occur due to their different sensor platforms, satellite positioning shifts, or acquisition conditions, as shown in Fig.~\ref{fig:challenge}; (iii) Temporal discrepancies occur due to different acquisition times, also resulting in the spatial misalignment with differences in object appearance and seasons. This makes it very complicated the SET problem (Fig.~\ref{fig:challenge}); (iv) Finally, the scarcity of paired SAR-EO datasets makes it challenging for existing methods to learn the SET effectively, often leading to overfitting or unstable results.

To address these limitations, we \textit{firstly} propose a novel framework that leverages a \textit{pretrained} Latent Diffusion Model (LDM) \cite{SD, SD3} to improve SAR-to-EO image translation, which is denoted as Confidence Diffusion for SAR-to-EO Translation (C-DiffSET). The advantages of utilizing the pretrained LDM for SET tasks are as follows: (i) Since RGB bands of EO images share visual characteristics with natural images, we fine-tune the pretrained LDM---trained on large-scale natural image datasets---to transfer its representation power to the SET task. This addresses the issue of limited SAR-EO paired data, as our framework can leverage the pretrained knowledge from natural image distributions; (ii) The pretrained LDM operates in a 1/8-downsampled latent space, inherently alleviating spatially local misalignments caused by imperfect alignment processes; (iii) The variational auto-encoder (VAE) \cite{VAE} from LDM can also be applicable for SAR images. As shown in Fig.~\ref{fig:vae}, despite the presence of significant speckle noise in SAR data, the VAE encoder effectively embeds SAR images into the same latent space as EO images, leveraging its denoising capability. This enables the smooth transfer of SAR latents as conditioning information in the reverse diffusion process, ensuring the generation of EO outputs with accurate pixel-wise correspondence to the SAR inputs.

In addition to utilizing the pretrained LDM as a foundation model for our SET task, we introduce a confidence-guided diffusion (C-Diff) loss to handle temporal discrepancies. The temporal discrepancies, such as objects appearing in only one modality, can introduce artifacts and hallucinated content in the generated EO images. Since these discrepancies originate from real-world acquisition conditions that SAR and EO images are often obtained for same regions at different time instances, they cannot be explicitly corrected in the dataset itself during the training process. To mitigate this, the U-Net \cite{unet} in our framework predicts both each predicted noise and its corresponding confidence map, which quantifies the pixel-wise uncertainty in the predicted noise. The U-Net receives a channel-wise concatenated SAR and noisy EO features as input, allowing it to jointly model both contexts. This setup enables the confidence map to reflect the inherent uncertainty caused by temporal discrepancies, guiding the C-Diff loss to adaptively reduce penalties in regions where SAR-EO discrepancies likely occur. As a result, the generated EO images achieve high pixel-wise fidelity while minimizing artifacts and hallucinations. Our key contributions are summarized as:
\begin{itemize}
    \item We propose C-DiffSET, the \textit{first} framework to fine-tune a \textit{pretrained} LDM for SET tasks, effectively leveraging their learned representations to overcome the scarcity of SAR-EO image pairs. In the C-DiffSET, SAR and EO images are embedded into the same latent space to ensure pixel-wise correspondence between SAR and EO latent throughout the framework;
    \item We introduce a novel C-Diff loss that can guide our C-DiffSET to reliably predict both EO outputs and confidence maps for accurate SET to mistigate locally mismatching challenges from temporal discrepancy due to their different acquisition times;
    \item We validate our C-DiffSET through extensive experiments on datasets with varying resolutions and ground sample distances (GSD) \cite{GSD}, including QXS-SAROPT \cite{QXS_SAROPT}, SAR2Opt \cite{SAROpt}, and SpaceNet6 \cite{SpaceNet6} datasets, demonstrating the superiority of our C-DiffSET that \textit{significantly} outperforms the very recent image-to-image translation methods and SET methods with \textit{large} margins.
\end{itemize}
\section{Related Work}
\label{sec:related}

\subsection{Image-to-Image Translation} Image-to-image translation has been widely studied in various fields, such as image colorization \cite{zhang2016colorful, zhang2017real}, style transfer \cite{CycleGAN, pix2pix, ASAP}, and image inpainting \cite{iizuka2017globally, pathak2016context}. These models typically rely on carefully constructed paired datasets across domains for effective training. For instance, Pix2Pix \cite{pix2pix} introduced conditional GANs with an $l_1$ loss for domain-specific translations. However, acquiring paired datasets remains challenging, and various methods have been proposed to work around unpaired data constraints \cite{baek2021rethinking, liu2021smoothing, kim2022instaformer, choi2018stargan, huang2018multimodal, nicenice}. CycleGAN \cite{CycleGAN} addressed this limitation using a cycle-consistency loss to facilitate training without paired data. Further addressing non-bijective translation, StegoGAN \cite{StegoGAN} introduced steganography into GAN-based models, enhancing semantic consistency in cases of domain mismatch by reducing spurious features in generated images without additional supervision.

\noindent\textbf{Diffusion-based approaches.}  
Diffusion models \cite{DDPM, SD} have gained significant traction in image-to-image translation due to their ability to model complex data distributions through iterative denoising. Traditional Denoising Diffusion Probabilistic Models (DDPMs) \cite{DDPM} perform diffusion directly in the pixel space, but their high computational cost limits scalability in high-resolution tasks. To address this, Latent Diffusion Models (LDMs) \cite{SD} move the diffusion process to a learned latent space, significantly reducing memory and computational overhead while preserving high-resolution details. Palette \cite{Palette} employs conditional DDPMs in the pixel domain for high-quality translations, whereas BBDM \cite{BBDM} performs diffusion directly in the latent space, eliminating the need for separate noise conditioning but imposing strict pixel-wise alignment requirements. DGDM \cite{DGDM} extends BBDM with a deterministic translator network to generate target features before diffusion, adding computational overhead and dependency on the translator’s performance. ControlNet \cite{ControlNet} and its variants, Uni-ControlNet \cite{UniControlNet}, extend pretrained diffusion models by injecting control signals through additional conditioning networks. These methods excel at preserving structural guidance for simple condition images, such as edge maps or skeleton poses. However, they are less suitable for the SET task, where the relationship between SAR and EO images is highly complex and requires learning intricate domain mappings rather than simple structural constraints.

\begin{figure*}[tbp]
  \centering
  \includegraphics[width=1.0\textwidth]{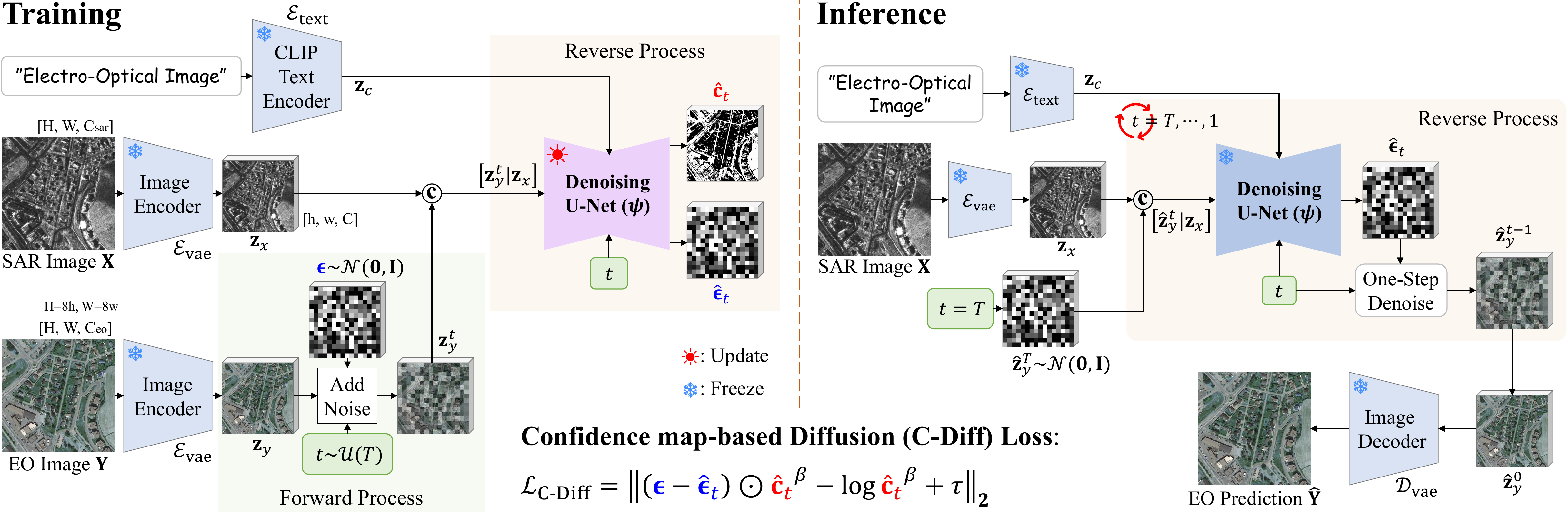}
  \vspace{-0.6cm}
  \caption{Overall framework of our Confidence Diffusion for SAR-to-EO Translation (C-DiffSET).}
  \label{fig:main}
  \vspace{-0.2cm}
\end{figure*}

\subsection{SAR-to-EO Image Translation (SET)}
Deep learning-based image-to-image translation methods \cite{pix2pix, ASAP, iizuka2017globally, huang2018multimodal, baek2021rethinking, liu2021smoothing, zheng2021spatially, BBDM} have been widely adopted to remote sensing applications, including SAR-to-EO image translation (SET). However, SAR imagery poses unique challenges due to inherent characteristics like speckle noise from backscattering effects \cite{goodman, speckle_noise, character_zhang, character_soergel, character_spigai, character_zhang2, character_huang1}, complicating the translation process. To address these challenges, most SET studies employ GAN-based approaches \cite{despeckle, cabrera, CFRWD, SGCLSET, kong, kento, zhang, youk2023transformer, MCGAN, nicegan, sebasti}. Conditional GAN-based models are commonly used to leverage noisy SAR information as a conditioning input, while CycleGAN-based models align the SAR and EO domains by enforcing cycle-consistency constraints. SAR-SMTNet \cite{youk2023transformer} introduced a Swin-Transformer-based \cite{liu2021swin} generator to improve structural consistency in SAR-to-EO translation. Additionally, CFCA-SET \cite{CFCASET} proposed a coarse-to-fine SAR-to-EO model that incorporates near-infrared (NIR) images during training to refine EO generation.

\noindent\textbf{Diffusion-based approaches.} Recently, a limited number of diffusion-based studies have been explored for SET, primarily divided into DDPM-based \cite{sar_ddpm1, sar_ddpm2} and LDM-based \cite{cBBDM, DSE, CMdiffusion} methods. DSE \cite{DSE} utilize the BBDM \cite{BBDM} to generate EO predictions for downstream tasks like flood segmentation. Expanding on this, cBBDM \cite{cBBDM} integrates SAR inputs as conditional information in BBDM to improve translation quality, while CM-Diffusion \cite{CMdiffusion} further enhances BBDM by conditioning on color features, helping to preserve spectral consistency in EO generation.

Despite these advancements, both GAN-based and diffusion-based methods encounter limitations stemming from the scarcity of paired SAR-EO datasets. The GAN-based models frequently face with convergence challenges and mode collapse under limited data, while diffusion-based approaches struggle with overfitting and slow convergence due to extensive training from scratch. Recent open-source \cite{diffusers}  releases of foundation LDMs, such as Stable Diffusion \cite{SD, SD3} and SDXL \cite{SDXL}, pretrained on large-scale text-to-image datasets (e.g., LAION-5B \cite{LAION}), have demonstrated strong generative priors for image synthesis. These models have been widely adapted for natural image applications \cite{wu2024seesr, zabari2023diffusing, ke2024repurposing, he2024lotus}, showcasing their flexibility in domain adaptation. Building on this foundation, we fine-tune a pretrained LDM for the SET task. By using the extensive visual representations learned from large-scale generative models, C-DiffSET effectively adapts to SAR-EO data, achieving robust translation while mitigating the risk of overfitting.
\section{Methods}
\label{sec:method}
\subsection{Overview of Proposed C-DiffSET}
We utilize a paired SAR-EO dataset $\mathcal{I} = \{(\mathbf{X}, \mathbf{Y})\}$, where $\mathbf{X} \in \mathbb{R}^{H \times W \times C_{\text{sar}}}$ represents SAR images of $H \times W$ sizes and $C_{\text{sar}}$ channels and $\mathbf{Y} \in \mathbb{R}^{H \times W \times C_{\text{eo}}}$ denotes EO images of $H \times W$ sizes and $C_{\text{eo}}$ channels. Our goal is to generate a predicted EO image $\widehat{\mathbf{Y}}$ corresponding to the given SAR input $\mathbf{X}$. Fig.~\ref{fig:main} illustrates the proposed Confidence Diffusion for SAR-to-EO Translation (C-DiffSET) framework, which comprises three key components: (i) the embedding of SAR and EO images into the latent space, (ii) the forward diffusion process, and (iii) the reverse diffusion process with the confidence-guided diffusion (C-Diff) loss. As described in Sec.~\ref{sec:32}, SAR image $\mathbf{X}$ and EO image $\mathbf{Y}$ are passed through the VAE encoder $\mathcal{E}_{\text{vae}}$, generating the latent features $\mathbf{z}_{x} \in \mathbb{R}^{h \times w \times C}$ (SAR feature) and $\mathbf{z}_{y} \in \mathbb{R}^{h \times w \times C}$ (EO feature). In the forward diffusion process (Sec.~\ref{sec:33}), noise $\bm{\epsilon}$ is added to the EO feature $\mathbf{z}_{y}$ over timesteps $t$s. During the reverse diffusion process, the U-Net $\psi$ takes the noisy EO feature $\mathbf{z}_{y}^{t}$, the timestep $t$, and the SAR feature $\mathbf{z}_{x}$ as conditional inputs. The Denoising U-Net $\psi$ predicts both the noise $\hat{\bm{\epsilon}}_t$ and a confidence map $\hat{\mathbf{c}}_t$ that quantifies pixel-wise uncertainty of the prediction. Finally, in Sec.~\ref{sec:34}, we describe the inference stage, where the predicted EO image $\widehat{\mathbf{Y}}$ is generated by reversing the noise-added process.

\subsection{SAR and EO Latent Space Generation}
\label{sec:32}
\begin{figure}[tbp]
  \centering
  \includegraphics[width=1.0\columnwidth]{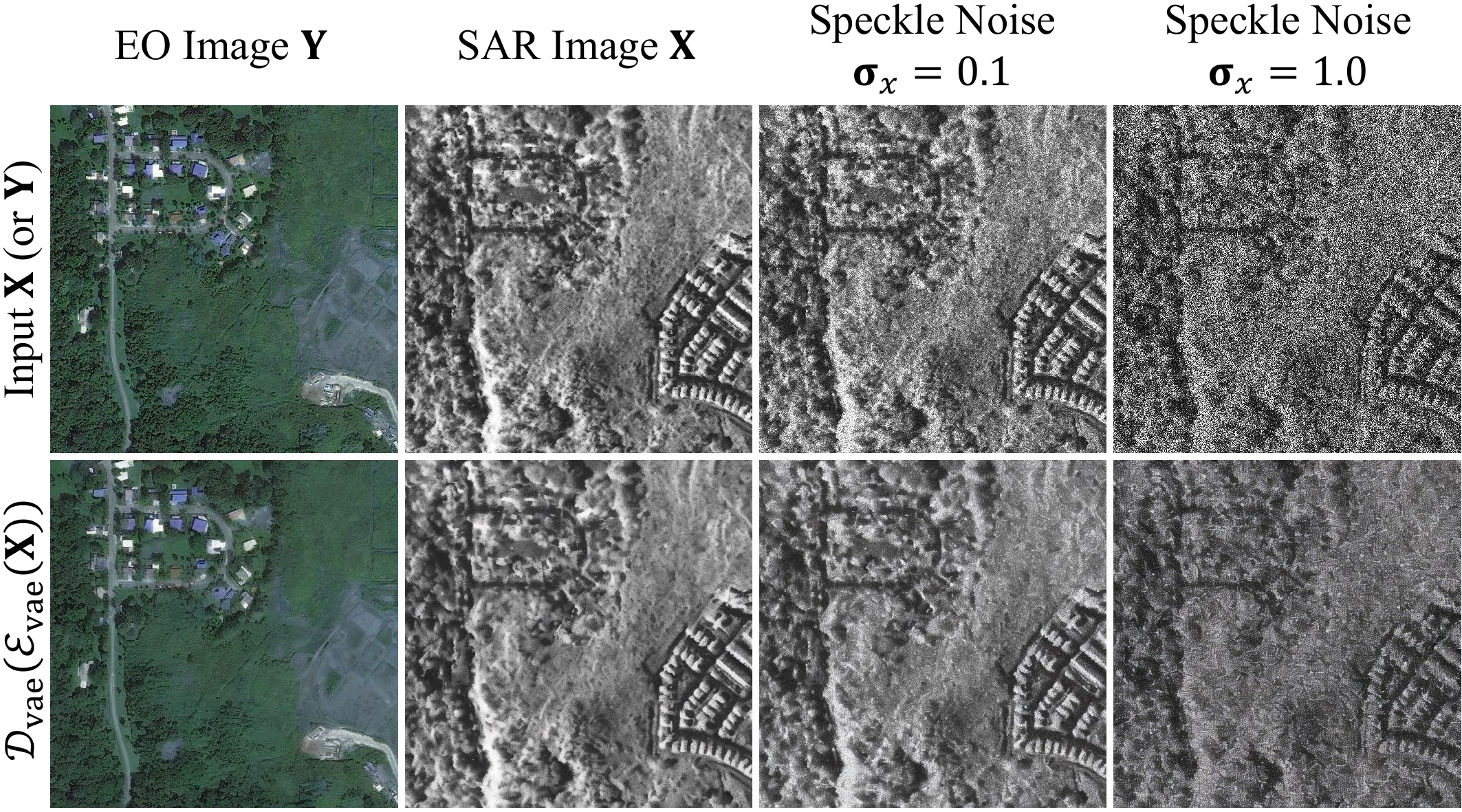}
  \caption{Results of applying the VAE encoder and decoder from LDM to EO and SAR images. The first row shows input images, including EO and SAR images with different levels of speckle noise. The second row presents the corresponding VAE reconstructions, illustrating that both EO and SAR images are accurately reconstructed despite noise variations.}
  \label{fig:vae}
\end{figure}

We utilize the pretrained VAE from LDM for mapping input images from the pixel space to the latent space. The VAE is frozen during our experiments and serves as both an encoder $\mathcal{E}_{\text{vae}}$ and a decoder $\mathcal{D}_{\text{vae}}$ for input images. Since the LDM has been pretrained on large-scale natural image datasets, the VAE is designed to accept 3-channel RGB images as input. Fig.~\ref{fig:vae} shows the results of applying the VAE encoder $\mathcal{E}_{\text{vae}}$ and decoder $\mathcal{D}_{\text{vae}}$ to SAR and EO images, validating its applicability for both modalities.

\noindent\textbf{EO latent space.} Since the RGB bands of EO images $\mathbf{Y}$ are naturally represented as 3-channel inputs, they are directly passed through the VAE without modification. In our experiments, we confirmed that the reconstruction error $\|\mathbf{Y}-\mathcal{D}_\text{vae}\left(\mathcal{E}_\text{vae}\left(\mathbf{Y}\right)\right)\|_{2}$ is minimal, ensuring that the VAE accurately preserves the content and structure of EO images. Furthermore, this low reconstruction error indicates that the RGB bands of EO images exhibit minimal domain gap compared to natural RGB images, making them well-suited for adaptation using pretrained LDMs. This indicates that the VAE’s output provides an upper bound on the achievable performance of our framework, serving as a stable baseline for EO image generation.

\noindent\textbf{SAR latent space.} SAR data $\mathbf{X}$, depending on the satellite sensors, is available as single-polarization SAR images of 1-channel (HH or VV component) or full-polarization ones of 4-channel (HH, HV, VH, and VV components). For 1-channel SAR images, we repeat the channel three times to match the VAE’s input requirements. For 4-channel SAR data, we construct a 3-channel input by using HH, the average of HV and VH, and VV components. After passing these inputs through the VAE, we observed that the reconstruction error $\|\mathbf{X}-\mathcal{D}_\text{vae}\left(\mathcal{E}_\text{vae}\left(\mathbf{X}\right)\right)\|_{2}$ remained low, and the outputs appeared visually pleasing, as shown in Fig.~\ref{fig:vae}. Additionally, we examined the VAE’s behavior under varying levels of speckle noise added to SAR inputs. The results demonstrate that the VAE’s reconstruction process adapts spatially to the noise level, effectively denoising the inputs while preserving structural details. This indicates that the VAE can embed SAR images into the same latent space as EO images without additional training, allowing SAR features to serve as conditioning inputs during the diffusion process. This ensures that pixel-wise correspondence between SAR and EO latent spaces is maintained throughout the generation process.

\subsection{Training Strategy for the Diffusion Process}
\label{sec:33}

The diffusion process consists of two key stages: a forward process that progressively adds noise to the target EO feature, and a reverse process that Denoising U-Net $\psi$ predicts and removes the added noise.

\noindent\textbf{Forward process.}  
Following the DDPMs \cite{DDPM} framework, noise is added to the target EO feature $\mathbf{z}_y$ over a sequence of timesteps $t\sim\mathcal{U}(T)$, where $\mathcal{U}$ is a uniform distribution and $T$ is a total number of timesteps. Specifically, at each timestep $t$, a noisy version of the EO feature $\mathbf{z}_y^t$ is generated by sampling:
\begin{equation}
\mathbf{z}_y^t = \sqrt{\bar{\alpha}_t} \mathbf{z}_y + \sqrt{1 - \bar{\alpha}_t} \bm{\epsilon},  \bm{\epsilon} \sim \mathcal{N}(\mathbf{0}, \mathbf{I}),
\end{equation}
where $\mathcal{N}$ is a Gaussian distribution, $\bm{\epsilon} \in \mathbb{R}^{h \times w \times C}$, and $\bar{\alpha}_{t} = \prod_{s=1}^{t} (1 - \beta_{s})$ \cite{DDPM} determines the noise magnitude at each timestep $t$.

\noindent\textbf{Reverse process.} $\psi$ learns the reverse process by predicting the added noise $\bm{\epsilon}$, aligning with the pretrained LDM design. To leverage the pre-trained LDM’s text-to-image capability and provide a strong initialization, we use a fixed prompt, $p=\text{``electro-optical image"}$, as a stable conditioning signal. This guides the LDM to focus on EO-specific features by embedding the prompt via the CLIP \cite{CLIP} text encoder $\mathcal{E}_{\text{text}}$ as $\mathbf{z}_c = \mathcal{E}_{\text{text}}(p)$, rather than using a null prompt. $\psi$ takes as input the noisy EO feature $\mathbf{z}_y^t$ and conditioned SAR feature $\mathbf{z}_x$, concatenated along the channel dimension, and is also fed with the timestep $t$ and the text prompt embedding $\mathbf{z}_c$. Then, we have two output components: a predicted noise $\hat{\bm{\epsilon}}_t$ and a confidence map $\hat{\mathbf{c}}_t \in \mathbb{R}^{h \times w \times 1}$ that contains the pixel-wise uncertainty of $\hat{\bm{\epsilon}}_t$:
\begin{equation}
[\hat{\bm{\epsilon}}_t \mid \hat{\mathbf{c}}_t] = \psi\left([\mathbf{z}_y^t \mid \mathbf{z}_x],\;\mathbf{z}_c,\;t\right),
\end{equation}
where $\left[\;\cdot\mid\cdot\;\right]$ indicates channel-wise concatenation. The confidence map $\hat{\mathbf{c}}_t$ is further processed through a $\mathsf{SoftPlus}$ \cite{softplus} operation to ensure all values remain non-negative.

\noindent\textbf{Confidence-guided diffusion (C-Diff) loss.} Seitzer \textit{et al.} \cite{seitzer2022pitfalls} proposed the $\beta$-NLL loss to capture aleatoric uncertainty for regression, classification and generative tasks. Inspired by the $\beta$-NLL loss \cite{SeaRaft, seitzer2022pitfalls}, we firstly adopt it into the training stage of diffusion process to enhance pixel-wise fidelity and to manage temporal inconsistencies between SAR and EO images.
We apply the confidence output of $\psi$ at each diffusion step to the $\beta$-NLL loss for stable SET, which is called C-Diff loss $\mathcal{L}_{\text{C-Diff}}$. That is, $\mathcal{L}_{\text{C-Diff}}$ uses a learned confidence map $\hat{\mathbf{c}}_t$ to adaptively weight the predicted noise $\hat{\bm\epsilon}_t$ pixel-wise. $\mathcal{L}_{\text{C-Diff}}$ is designed to prioritize high-confidence areas while deweighting uncertain regions, thus improving robustness in regions with temporal misalignments (e.g., dynamic objects in SAR or EO images). $\mathcal{L}_{\text{C-Diff}}$ optimizes $\psi$ by combining a weighted pixel-wise reconstruction loss and a regularization term for the confidence map $\hat{\mathbf{c}}_t$:
\begin{equation}
    \mathcal{L}_{\text{C-Diff}} = \left\lVert\left(\bm{\epsilon} - \hat{\bm{\epsilon}}_t\right)\odot{\hat{\mathbf{c}}_t}^{\beta} - \log {\hat{\mathbf{c}}_t}^{\beta} + \tau\right\rVert_2,
\end{equation}
where $\tau$ is a margin term ensuring non-negativity of the loss. $\hat{\mathbf{c}}_t$ effectively acts as an adaptive weighting factor, allowing the model to focus more on well-aligned regions and reduce penalties in uncertain areas. The $\log$ term serves as a regularizer, preventing $\hat{\mathbf{c}}_t$ from collapsing to zero. Empirically, we found that setting $\beta=1$ yields the best performance, while $\beta=0$ reduces $\mathcal{L}_{\text{C-Diff}}$ to a standard $\ell_2$ (MSE) loss, lacking adaptive weighting. $\mathcal{L}_{\text{C-Diff}}$ allows C-DiffSET to produce EO outputs of high structural accuracy while mitigating artifacts and hallucinations that can arise in temporally inconsistent regions. 

\subsection{Inference Stage for EO Image Prediction}
\label{sec:34}
Given an unseen SAR image $\mathbf{X}$, we pass it through the VAE encoder $\mathcal{E}_{\text{vae}}$ to obtain its SAR latent code $\mathbf{z}_{x}=\mathcal{E}_{\text{vae}}(\mathbf{X})$, which serves as the conditioning information for the reverse diffusion process. For inference, the Denoising U-Net $\psi$ iteratively refines the noisy latent code $\hat{\mathbf{z}}_{y}^{T}$, to reconstruct the target EO latent code $\mathbf{z}_{y}^{0}=\mathbf{z}_{y}$ by denoising it back to the clean EO latent code through noise prediction $\hat{\bm{\epsilon}}_t$. Starting from pure noise, $\hat{\mathbf{z}}_{y}^{T} \sim \mathcal{N}(\mathbf{0}, \mathbf{I})$, $\psi$ iteratively refines the EO latent code $\hat{\mathbf{z}}_{y}^{t}$ by predicting the noise $\hat{\bm{\epsilon}}_t$ to be removed at each timestep $t$ as:
\begin{equation}
    [\hat{\bm{\epsilon}}_t \mid \mathsf{Dummy}] = \psi\left([\hat{\mathbf{z}}_y^t \mid \mathbf{z}_x],\;\mathbf{z}_c,\;t\right),
\end{equation}
where $\mathsf{Dummy}$ indicates dummy confidence values. The predicted noise $\hat{\bm\epsilon}_t$ is then used to compute $\hat{\mathbf{z}}_{y}^{t-1}$ in the reverse diffusion process, following the formulation in \cite{DDPM}, which progressively denoises the EO latent code until it converges to the target EO latent code. The final EO latent code $\hat{\mathbf{z}}_{y}^{0}$ is passed through the VAE decoder $\mathcal{D}_{\text{vae}}$ to generate the reconstructed EO image $\widehat{\mathbf{Y}} = \mathcal{D}_{\text{vae}} (\hat{\mathbf{z}}_{y}^{0})$. 

\begin{figure*}[tbp]
  \centering
  \includegraphics[width=1.0\textwidth]{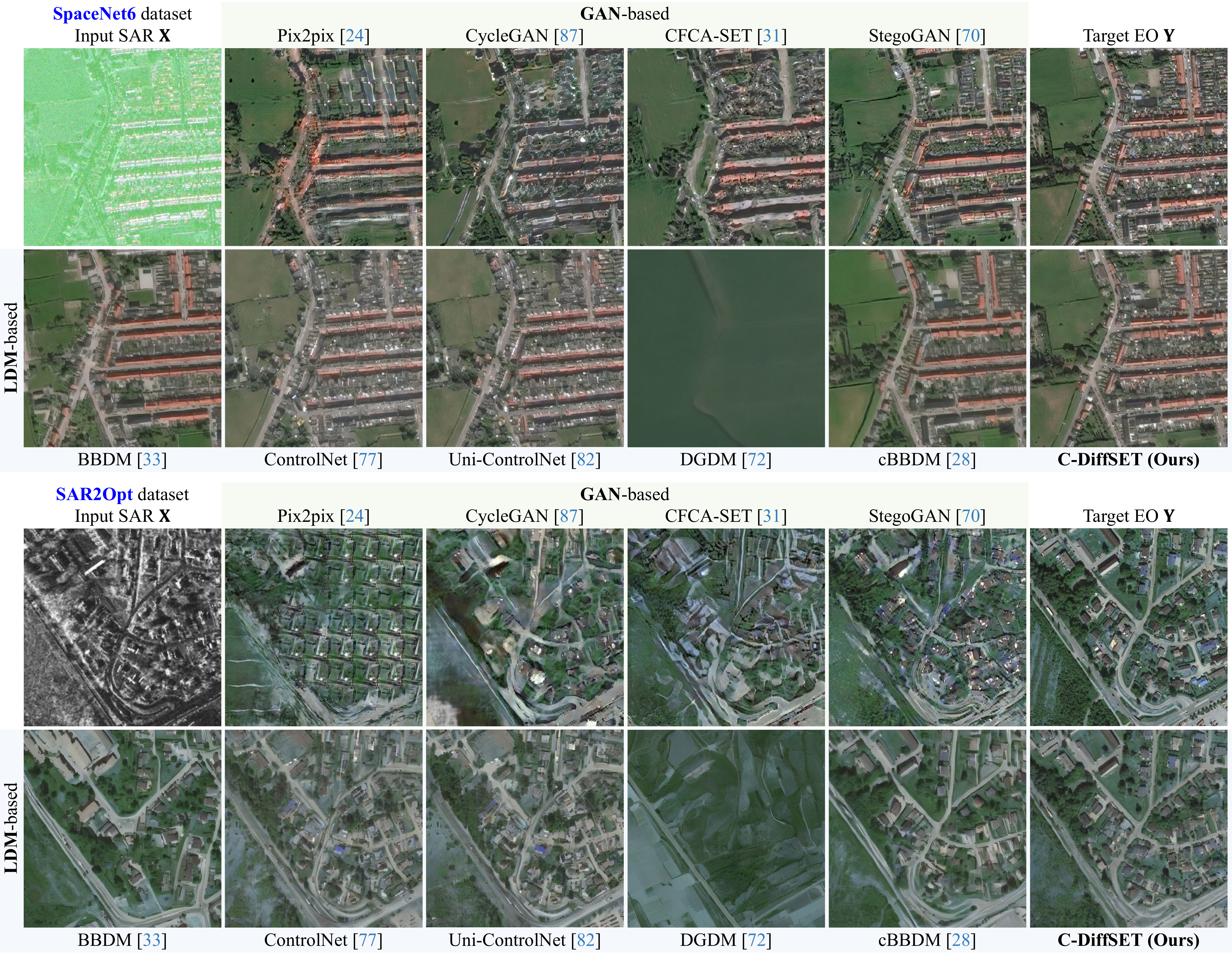}
  \vspace{-0.5cm}
  \caption{Visual comparison of SET results on the SpaceNet6 and SAR2Opt datasets. 1st rows: GAN-based (Pix2pix, CycleGAN, CFCA-SET, and StegoGAN) methods. 2nd rows: LDM-based (BBDM, ControlNet, Uni-ControlNet, DGDM, cBBDM, and C-DiffSET) methods.}
  \label{fig:spacenet}
  \vspace{-0.3cm}
\end{figure*}

\section{Experiment}
\label{sec:experiment}

\subsection{Datasets}
\noindent We evaluate our C-DiffSET framework on three publicly available SAR-to-EO datasets: QXS-SAROPT \cite{QXS_SAROPT}, SAR2Opt \cite{SAROpt}, and SpaceNet6 \cite{SpaceNet6}. These datasets vary in satellite platforms, GSD \cite{GSD}, and SAR polarization modes, allowing us to assess the robustness and generalizability of our approach across diverse real-world scenarios. The SAR images in these publicaly released datasets are provided as \textit{magnitude-only} representation, containing real-valued intensity without phase information.

\noindent\textbf{QXS-SAROPT} \cite{QXS_SAROPT}.  
The QXS-SAROPT dataset contains 20,000 SAR and EO image pairs captured by the Gaofen-3 satellite (SAR) and Google Earth (EO). The SAR images are acquired in a single-polarization mode. The EO images consist of RGB channels, covering various port cities. Each image patch measures 256$\times$256 pixels with an 1-$m$ GSD, focusing on complex maritime environments.

\noindent\textbf{SAR2Opt} \cite{SAROpt}.  
The SAR2Opt dataset provides 2,076 SAR and EO image pairs obtained from the TerraSAR-X satellite (SAR) and Google Earth (EO). The SAR images are captured in a single-polarization mode. The corresponding EO images contain RGB channels and cover diverse Asian cities. Each patch measures 600$\times$600 pixels with an 1-$m$ GSD, making this dataset particularly useful for urban area analysis and infrastructure monitoring.

\noindent\textbf{SpaceNet6} \cite{SpaceNet6}.  
The SpaceNet6 dataset offers SAR and EO image pairs captured by Capella Space (SAR) and Maxar WorldView-2 (EO) satellites. The SAR images are acquired with full-polarization, enabling detailed analysis of surface structures. The EO images include RGB and NIR bands, though only the RGB subset is used in our experiments. This dataset contains 3,401 SAR-EO image pairs, with each patch size of 900$\times$900 pixels and a 0.5-$m$ GSD, focusing on urban landscapes and building detection tasks.

\subsection{Experiment Details}
All experiments were implemented using PyTorch \cite{Pytorch} and conducted on a single NVIDIA A6000 GPU. Each model was fine-tuned for 50,000 iterations, with a 100-step warmup period. We employed the AdamW optimizer \cite{adamw} with an initial learning rate of $3 \times 10^{-5}$ and a weight decay of 0.01. A cosine-annealing scheduler \cite{CosineAnneal} was used to progressively reduce the learning rate at each iteration. For the pretrained LDM, we utilized Stable Diffusion v2.1 \cite{SD, diffusers}, and the text encoder was frozen as the CLIP-ViT-H/14 \cite{CLIP, openclip} text encoder. For the latent features, we set the spatial resolution to $h = H/8$ and $w = W/8$, and the channel dimension to $C = 4$. The training noise scheduler was based on DDPM \cite{DDPM} with a total of 1,000 steps. For inference, we employed an efficient DDIM \cite{DDIM} noise scheduler with total 50 inference steps to accelerate the generation process. We evaluated the performance of our framework using Fréchet Inception Distance (FID) \cite{FID}, Learned Perceptual Image Patch Similarity (LPIPS) \cite{LPIPS}, Spatial Correlation Coefficient (SCC) \cite{scc}, Structural Similarity Index (SSIM) \cite{ssim}, and Peak Signal-to-Noise Ratio (PSNR).

\subsection{Experimental Results}
For comparative analysis, we used official implementations for general image-to-image translation methods \cite{pix2pix, CycleGAN, StegoGAN, BBDM, DGDM, ControlNet, UniControlNet}. For LDM-based methods, including our C-DiffSET, it should be noted that we ensured fair comparison by initializing all models with the same Stable Diffusion v2.1 weights (VAE, U-Net). For SET-specific methods \cite{CFCASET, cBBDM, youk2023transformer} (marked by $\dagger$), where their official codes are often unavailable due to this specific field, we re-implemented each method according to their technical descriptions.

\noindent\textbf{Qualitative comparison.} As shown in Fig.~\ref{fig:first} and Fig.~\ref{fig:spacenet}, the GAN-based methods generally struggle with stability during training, leading to prominent artifacts in SET. The LDM-based methods, such as BBDM \cite{BBDM} and cBBDM \cite{cBBDM}, face challenges due to their reliance on direct diffusion from SAR input $\mathbf{X}$ to EO output $\mathbf{Y}$. This setup makes them particularly susceptible to local spatial misalignments, producing blurred and incoherent results. DGDM \cite{DGDM} seeks to improve the initial stage of diffusion through a deterministic approach, employing a lightweight translation network to create a translated EO latent from SAR latent. However, the simplistic nature of this translation network fails to capture the intricate EO features, resulting in suboptimal initial latents for the diffusion process. Furthermore, ControlNet-based approaches \cite{ControlNet, UniControlNet} keep the pretrained U-Net weights frozen and only introduce SAR conditions at the decoder stage, limiting their ability to fully integrate SAR structural information. This constraint results in weaker feature adaptation and reduced robustness to SAR-induced artifacts. In contrast, our C-DiffSET leverages pretrained LDM as foundational model, effectively addressing alignment issues via a confidence-guided diffusion loss, leading to yield sharper and more structurally coherent EO images, and to outperform both GAN-based and diffusion-based baselines in visual fidelity. 

\noindent\textbf{Quantitative evaluation.}  
In Table~\ref{tab:main} and Table~\ref{tab:qxs}, the GAN-based methods benefit from the richer polarization diversity in SpaceNet6 dataset (4-channel full-polarization SAR data), yielding higher SSIM and SCC scores compared to 1-channel single-polarization datasets. However, for the other peceptual quality metrics such as FID and LPIPS, the GAN-based methods yield lower performance across datasets, indicating their inherent limitations in handling the SET task. The LDM-based methods, BBDM and cBBDM, also perform inadequately due to their direct diffusion setup from SAR to EO, which exacerbates the pixel-wise misalignment problem and inflates LPIPS and FID scores. The DGDM incorporates a lightweight translator network to initialize the diffusion process, but its simplistic structure fails to encapsulate detailed EO features, resulting in lower PSNR and SSIM scores. In contrast, our C-DiffSET achieves superior performance across all metrics, enabled by the pretrained LDM and confidence-guided diffusion loss, which jointly tackle SAR-specific noise and alignment challenges. This adaptation allows C-DiffSET to attain the highest PSNR, SSIM, and SCC values, along with the lowest LPIPS and FID scores, highlighting its enhanced fidelity and perceptual quality in SET.

\begin{table*}[tbp]
    \scriptsize
    \centering
    \resizebox{1.0\textwidth}{!}{
    \def\arraystretch{1.2}
    \begin{tabular} {c|l|l|c|c|c|c|c|c|c|c|c|c}
        \Xhline{2\arrayrulewidth}
        \multirow{2}{*}{Types} & \multirow{2}{*}{Methods} & \multirow{2}{*}{Publications} &\multicolumn{5}{c|}{\textbf{SAR2Opt} Dataset} & \multicolumn{5}{c}{\textbf{SpaceNet6} Dataset} \\
        \cline{4-13}
        & & & FID$\downarrow$ & LPIPS$\downarrow$ & SCC$\uparrow$ & SSIM$\uparrow$ & PSNR$\uparrow$ & FID$\downarrow$ & LPIPS$\downarrow$ & SCC$\uparrow$ & SSIM$\uparrow$ & PSNR$\uparrow$ \\
        \hline
        \multirow{5}{*}{GANs} & Pix2Pix \cite{pix2pix} & CVPR 2017 & 196.87 & 0.426 & 0.0006 & 0.216 & 15.422 & 124.55 & 0.256 & 0.0102 & 0.522 & 19.357 \\
        & CycleGAN \cite{CycleGAN} & ICCV 2017 & 139.72 & 0.425 & 0.0022 & 0.224 & 14.931 & 114.81 & 0.274 & 0.0097 & 0.493 & 17.798 \\
        & SAR-SMTNet$^\dag$ \cite{youk2023transformer} & TGRS 2023 & 160.87 & 0.479 & 0.0011 & 0.219 & 14.661 & 118.96 & 0.294 & 0.0103 & 0.483 & 17.032 \\
        & CFCA-SET$^\dag$ \cite{CFCASET} & TGRS 2023 & 152.27 & 0.430 & 0.0009 & 0.223 & 15.183 & 164.78 & 0.279 & 0.0097 & 0.498 & 18.297 \\
        & StegoGAN \cite{StegoGAN} & CVPR 2024 & 144.54 & 0.398 & 0.0034 & 0.237 & 15.624 & 75.12 & 0.244 & 0.0106 & 0.516 & 18.958 \\
        \hline
        \multirow{6}{*}{LDMs} & BBDM \cite{BBDM} & CVPR 2023 & 94.72 & 0.473 & 0.0005 & 0.234 & 15.131 & 81.86 & 0.302 & 0.0019 & 0.217 & 17.678 \\
        & ControlNet \cite{ControlNet} & ICCV 2023 & 81.04 & 0.423 & 0.0005 & 0.216 & 14.461 & 106.59 & 0.392 & 0.0027 & 0.178 & 14.085 \\
        & Uni-ControlNet \cite{UniControlNet} & NeurIPS 2023 & 80.81 & 0.421 & 0.0004 & 0.215 & 14.384 & 91.14 & 0.321 & 0.0037 & 0.183 & 14.333 \\
        & DGDM \cite{DGDM} & ECCV 2024 & 156.12 & 0.541 & 0.0004 & 0.273 & 15.568 & 238.37 & 0.438 & 0.0015 & 0.253 & 17.124 \\
        & cBBDM$^\dag$ \cite{cBBDM} & arXiv 2024 & 97.64 & 0.394 & 0.0022 & 0.285 & 16.591 & 72.77 & 0.243 & 0.0079 & 0.254 & 19.033 \\
        & \textbf{C-DiffSET (Ours)} & - & {\color{red}{\textbf{77.81}}} & {\color{red}{\textbf{0.346}}} & {\color{red}{\textbf{0.0035}}} & {\color{red}{\textbf{0.286}}} & {\color{red}{\textbf{16.613}}} & {\color{red}{\textbf{37.44}}} & {\color{red}{\textbf{0.142}}} & {\color{red}{\textbf{0.0151}}} & {\color{red}{\textbf{0.567}}} & {\color{red}{\textbf{21.022}}} \\
        \Xhline{2\arrayrulewidth}
    \end{tabular}}
    \vspace{-0.2cm}
    \caption{Quantitative comparison of image-to-image translation methods and SET methods on SAR2Opt and SpaceNet6 datasets. \textbf{\color{red}{Red}} indicate the best performance in each metric.}
    \vspace{-0.2cm}
    \label{tab:main}
\end{table*}

\begin{table}[tbp]
    \scriptsize
    \centering
    \resizebox{1.0\columnwidth}{!}{
    \def\arraystretch{1.2}
    \begin{tabular} {c|l|c|c|c|c|c}
        \Xhline{2\arrayrulewidth}
        \multirow{2}{*}{Types} & \multirow{2}{*}{Methods} &\multicolumn{5}{c}{\textbf{QXS-SAROPT} Dataset} \\
        \cline{3-7}
        & & FID$\downarrow$ & LPIPS$\downarrow$ & SCC$\uparrow$ & SSIM$\uparrow$ & PSNR$\uparrow$ \\
        \hline
        \multirow{5}{*}{GANs} & Pix2Pix \cite{pix2pix} & 196.89 & 0.454 & 0.0000 & 0.247 & 14.924 \\
        & CycleGAN \cite{CycleGAN} & 195.38 & 0.455 & 0.0001 & 0.251 & 14.977 \\
        & SAR-SMTNet$^\dag$ \cite{youk2023transformer} & 117.69 & 0.435 & 0.0003 & 0.260 & 14.491 \\
        & CFCA-SET$^\dag$ \cite{CFCASET} & 79.06 & 0.406 & 0.0006 & 0.273 & 15.094 \\
        & StegoGAN \cite{StegoGAN} & 85.60 & 0.391 & 0.0019 & 0.280 & 15.580 \\
        \hline
        \multirow{6}{*}{LDMs} & BBDM \cite{BBDM} & 65.15 & 0.522 & 0.0004 & 0.238 & 13.946 \\
        & ControlNet \cite{ControlNet} & 22.39 & 0.434 & 0.0001 & 0.257 & 14.062 \\
        & Uni-ControlNet \cite{UniControlNet} & 22.48 & 0.437 & 0.0002 & 0.257 & 13.985 \\
        & DGDM \cite{DGDM} & 147.23 & 0.634 & 0.0001 & 0.288 & 11.564 \\
        & cBBDM$^\dag$ \cite{cBBDM} & 69.47 & 0.420 & 0.0023 & 0.304 & 16.248 \\
        & \textbf{C-DiffSET (Ours)} & {\color{red}{\textbf{18.15}}} & {\color{red}{\textbf{0.293}}} & {\color{red}{\textbf{0.0108}}} & {\color{red}{\textbf{0.372}}} & {\color{red}{\textbf{18.077}}} \\
        \Xhline{2\arrayrulewidth}
    \end{tabular}}
    \vspace{-0.2cm}
    \caption{Quantitative comparison of image-to-image translation methods and SET methods on the QXS-SAROPT dataset.}
    \vspace{-0.2cm}
    \label{tab:qxs}
\end{table}

\subsection{Ablation Studies}

\begin{figure}[tbp]
  \centering
  \includegraphics[width=1.0\columnwidth]{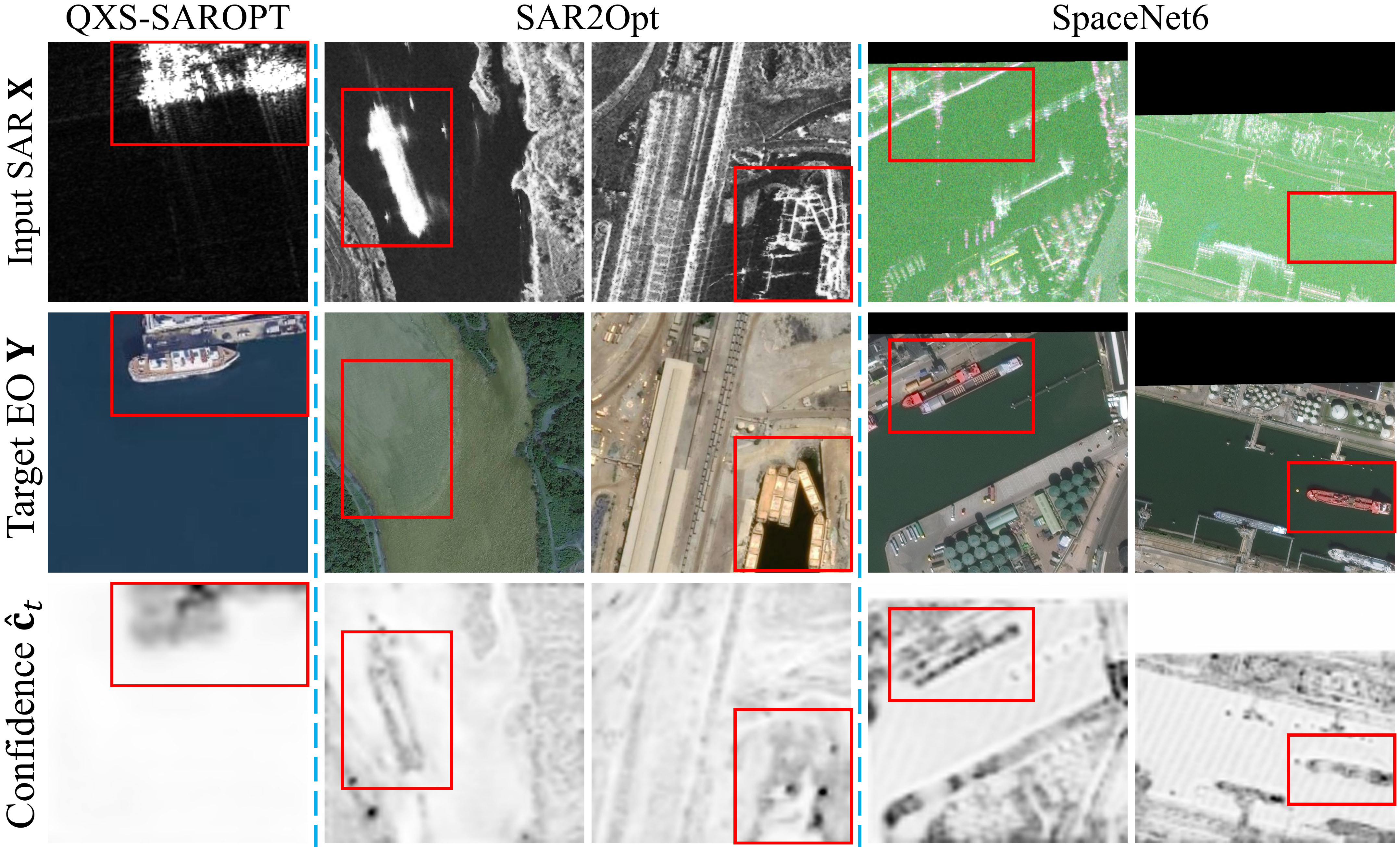}
  \vspace{-0.5cm}
  \caption{Confidence maps generated by C-DiffSET at timestep $t = T/2$ on SAR-EO paired datasets: QXS-SAROPT, SAR2Opt, and SpaceNet6. Each row illustrates the input SAR image $\mathbf{X}$, the target EO image $\mathbf{Y}$, and the corresponding confidence map $\hat{\mathbf{c}}_t$.}
  \label{fig:conf}
  \vspace{-0.3cm}
\end{figure}

\noindent\textbf{Impact of pretrained LDM and $\mathcal{L}_{\text{C-Diff}}$.} We ablate the contribution of the pretrained LDM and the confidence-guided diffusion loss $\mathcal{L}_{\text{C-Diff}}$ for the SAR2Opt and SpaceNet6 datasets, and show the results in Table~\ref{tab:ablation}. The pretrained LDM provides strong initialization for EO generation, improving the baseline PSNR and SSIM scores when compared to training from scratch. Even though the LDM-based methods and our C-DiffSET utilize the same pretrained LDM weights, our C-DiffSET outperformed the others by embedding conditioned SAR and target EO images into the shared latent space, leading to stable convergence. Furthermore, $\mathcal{L}_{\text{C-Diff}}$ helps enhancing the SSIM and PSNR metrics and lowering LPIPS and FID values, thus improving both perceptual fidelity and structural consistency. Fig.~\ref{fig:conf} shows confidence maps generated on the training dataset at timestep $t = T/2$ (midpoint of the denoising process), highlighting the areas of temporal discrepancy in SAR-EO pairs. This loss enables the U-Net $\psi$ to down-weight uncertain regions where objects are temporally misaligned across modalities, thereby reducing artifacts and ensuring coherent EO outputs. Further results and detailed analysis can be found in the \textit{Supplemental Material}.

\subsection{Limitations}

\noindent\textbf{Scalability across diverse satellites.} The VAE in C-DiffSET is designed for 3-channel inputs, matching typical RGB images. This limits its direct application to datasets with more channels (e.g., RGB+NIR or multispectral data). An extension to supporting such inputs would require further exploration and tuning for effective adaptation.

\begin{table}[tbp]
    \scriptsize
    \centering
    \resizebox{1.0\columnwidth}{!}{
    \def\arraystretch{1.2}
    \begin{tabular} {c|c|c|c|c|c|c}
        \Xhline{2\arrayrulewidth}
        \multirow{2}{*}{\makecell{Pretrained\\LDM}} & \multirow{2}{*}{\makecell{Loss\\function}} &\multicolumn{5}{c}{\textbf{SAR2Opt} / \textbf{SpacNet6}  Dataset} \\
        \cline{3-7}
        & & FID$\downarrow$ & LPIPS$\downarrow$ & SCC$\uparrow$ & SSIM$\uparrow$ & PSNR$\uparrow$ \\
        \hline
        & MSE & 98.98 / 60.26 & 0.39 / 0.23 & 0.001 / 0.011 & 0.26 / 0.43 & 15.82 / 18.16 \\
        \checkmark & MSE & 78.14 / 40.62 & 0.36 / 0.16 & 0.003 / 0.014 & 0.28 / 0.52 & 16.46 / 20.29 \\
        \checkmark& C-Diff& {\color{red}{\textbf{77.81}}} / {\color{red}{\textbf{37.44}}} & {\color{red}{\textbf{0.34}}} / {\color{red}{\textbf{0.14}}} & {\color{red}{\textbf{0.004}}} / {\color{red}{\textbf{0.015}}} & {\color{red}{\textbf{0.29}}} / {\color{red}{\textbf{0.57}}} & {\color{red}{\textbf{16.61}}} / {\color{red}{\textbf{21.02}}} \\
        \Xhline{2\arrayrulewidth}
    \end{tabular}}
    \vspace{-0.2cm}
    \caption{Ablation studies on the SAR2Opt and SpaceNet6 dataset evaluating the impact of pretrained LDM and confidence-guided diffusion (C-Diff) loss.}
    \label{tab:ablation}
\end{table}

\section{Conclusion}
\label{sec:conclusion}
In this work, we propose C-DiffSET, a novel framework that addresses key challenges in SAR-to-EO image translation. To the best of our knowledge, this is the first method to fully leverage LDM for SET tasks, mitigating issues caused by the limited availability of paired datasets. We introduce a C-Diff loss to handle temporal discrepancy between SAR and EO acquisitions, ensuring pixel-wise fidelity by adaptively suppressing artifacts and hallucinations. C-DiffSET achieves SOTA performance across datasets with varying GSD, demonstrating its effectiveness in real-world scenarios. Our framework provides a scalable foundation for future SET research and can be extended to other remote sensing applications and modalities.
\appendix

\section{Additional Discussions on Results}

\subsection{Additional Qualitative Comparisons}

\begin{figure*}[tbp]
  \centering
  \includegraphics[width=1.0\textwidth]{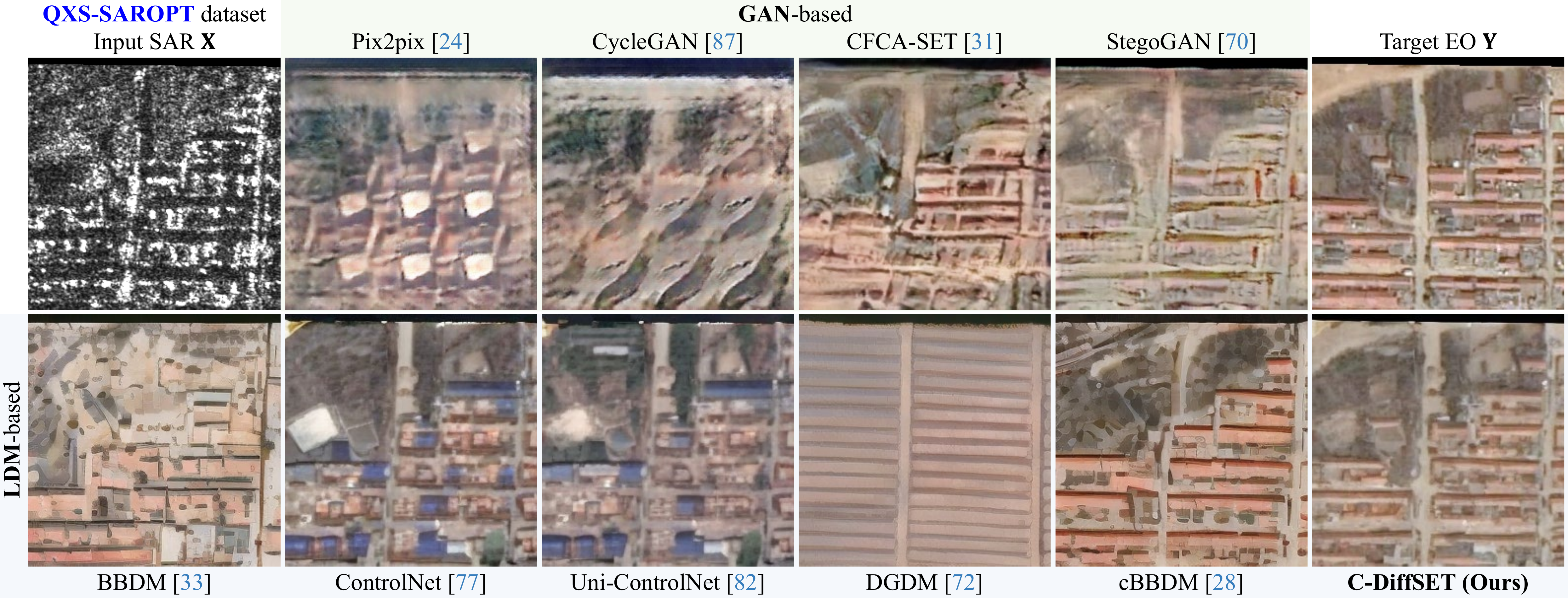}
  \caption{Visual comparison of SET results on the QXS-SAROPT dataset. 1st rows: GAN-based (Pix2pix, CycleGAN, CFCA-SET, and StegoGAN) methods. 2nd rows: LDM-based (BBDM, ControlNet, Uni-ControlNet, DGDM, cBBDM, and C-DiffSET) methods.}
  \label{fig:qxs_v1}
\end{figure*}

\begin{figure*}[tbp]
  \centering
  \includegraphics[width=1.0\textwidth]{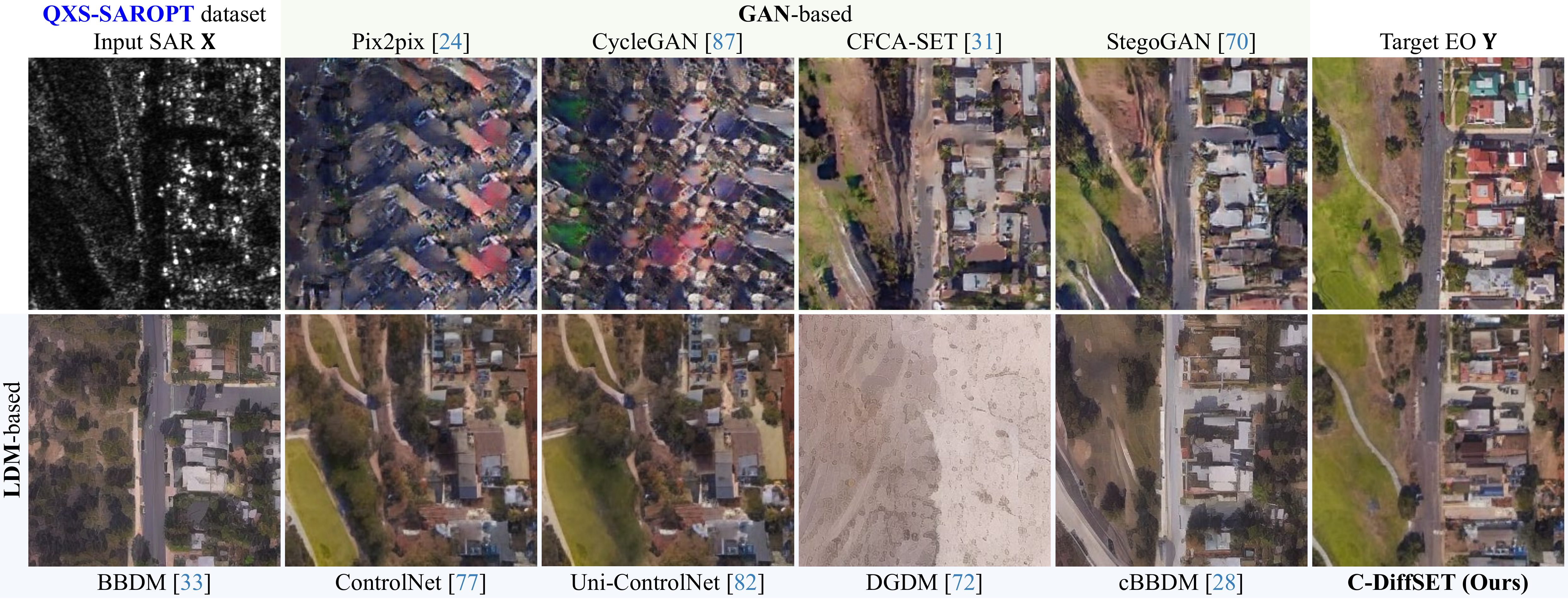}
  \caption{Visual comparison of SET results on the QXS-SAROPT dataset. 1st rows: GAN-based (Pix2pix, CycleGAN, CFCA-SET, and StegoGAN) methods. 2nd rows: LDM-based (BBDM, ControlNet, Uni-ControlNet, DGDM, cBBDM, and C-DiffSET) methods.}
  \label{fig:qxs_v2}
\end{figure*}

\begin{figure*}[tbp]
  \centering
  \includegraphics[width=1.0\textwidth]{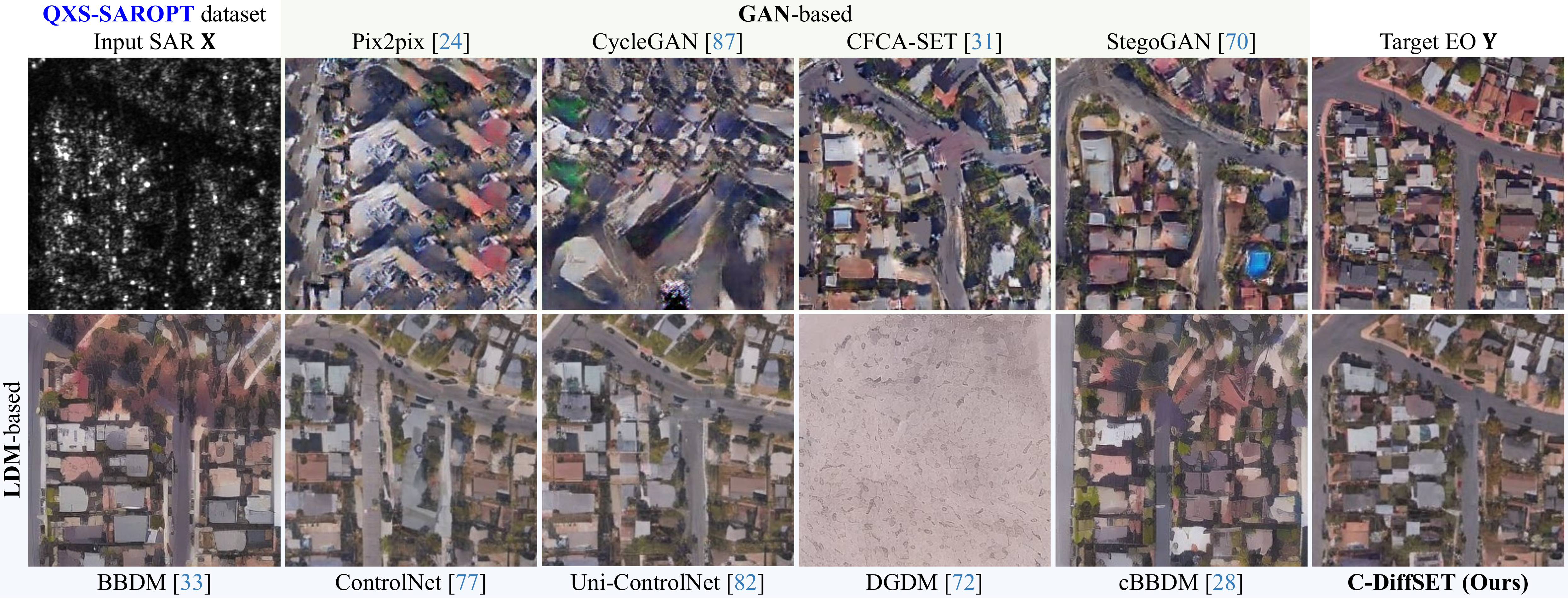}
  \caption{Visual comparison of SET results on the QXS-SAROPT dataset. 1st rows: GAN-based (Pix2pix, CycleGAN, CFCA-SET, and StegoGAN) methods. 2nd rows: LDM-based (BBDM, ControlNet, Uni-ControlNet, DGDM, cBBDM, and C-DiffSET) methods.}
  \label{fig:qxs_v3}
\end{figure*}

\begin{figure*}[tbp]
  \centering
  \includegraphics[width=1.0\textwidth]{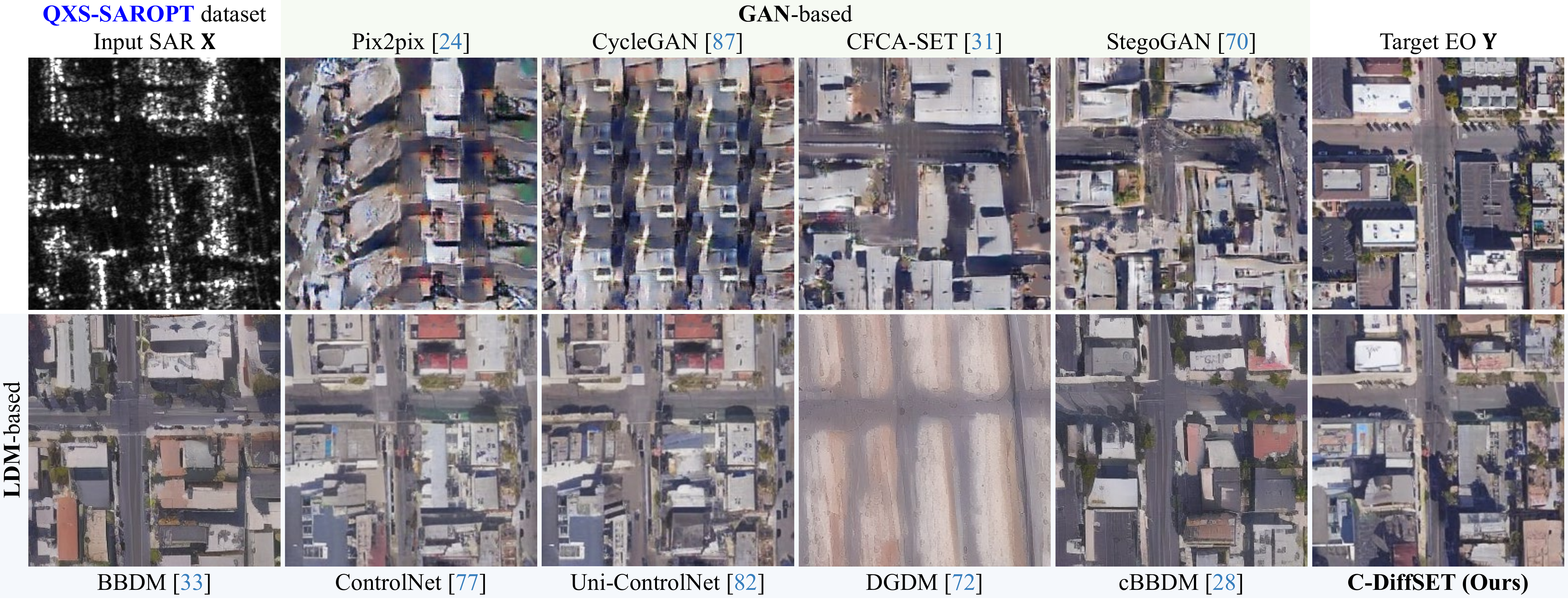}
  \caption{Visual comparison of SET results on the QXS-SAROPT dataset. 1st rows: GAN-based (Pix2pix, CycleGAN, CFCA-SET, and StegoGAN) methods. 2nd rows: LDM-based (BBDM, ControlNet, Uni-ControlNet, DGDM, cBBDM, and C-DiffSET) methods.}
  \label{fig:qxs_v4}
\end{figure*}

\begin{figure*}[tbp]
  \centering
  \includegraphics[width=1.0\textwidth]{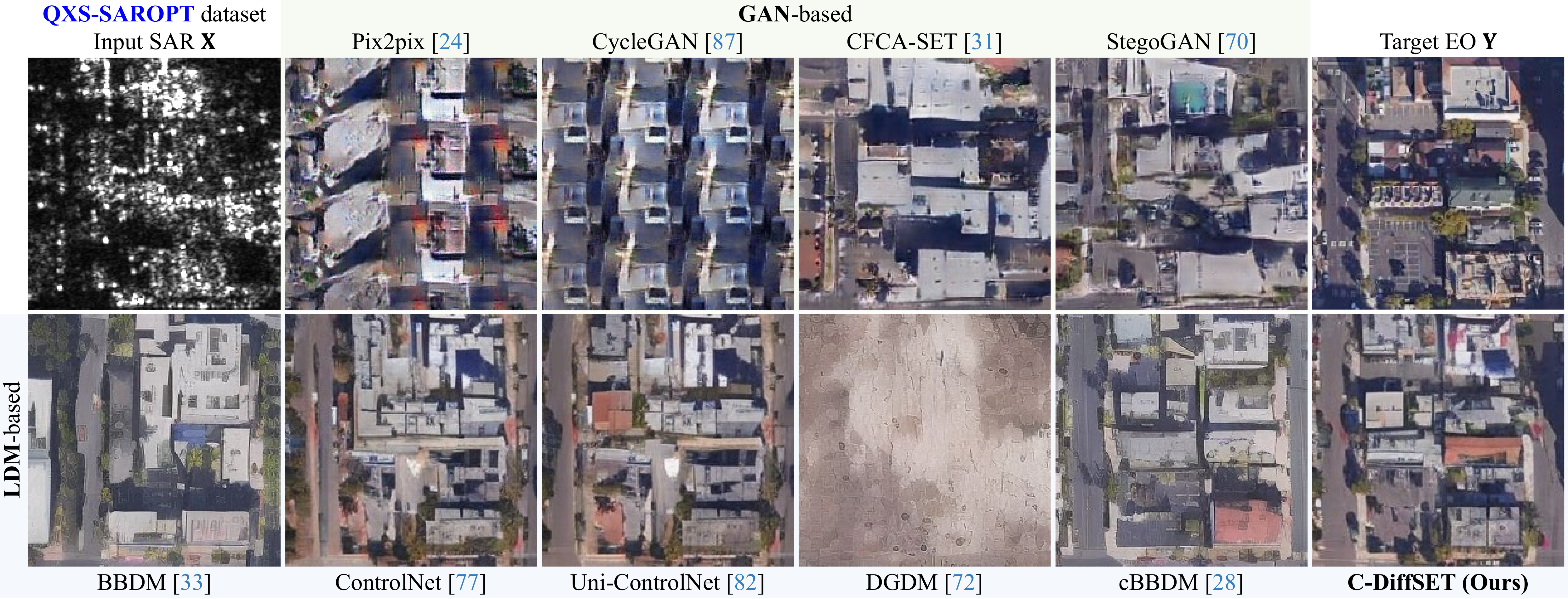}
  \caption{Visual comparison of SET results on the QXS-SAROPT dataset. 1st rows: GAN-based (Pix2pix, CycleGAN, CFCA-SET, and StegoGAN) methods. 2nd rows: LDM-based (BBDM, ControlNet, Uni-ControlNet, DGDM, cBBDM, and C-DiffSET) methods.}
  \label{fig:qxs_v5}
\end{figure*}

\begin{figure*}[tbp]
  \centering
  \includegraphics[width=1.0\textwidth]{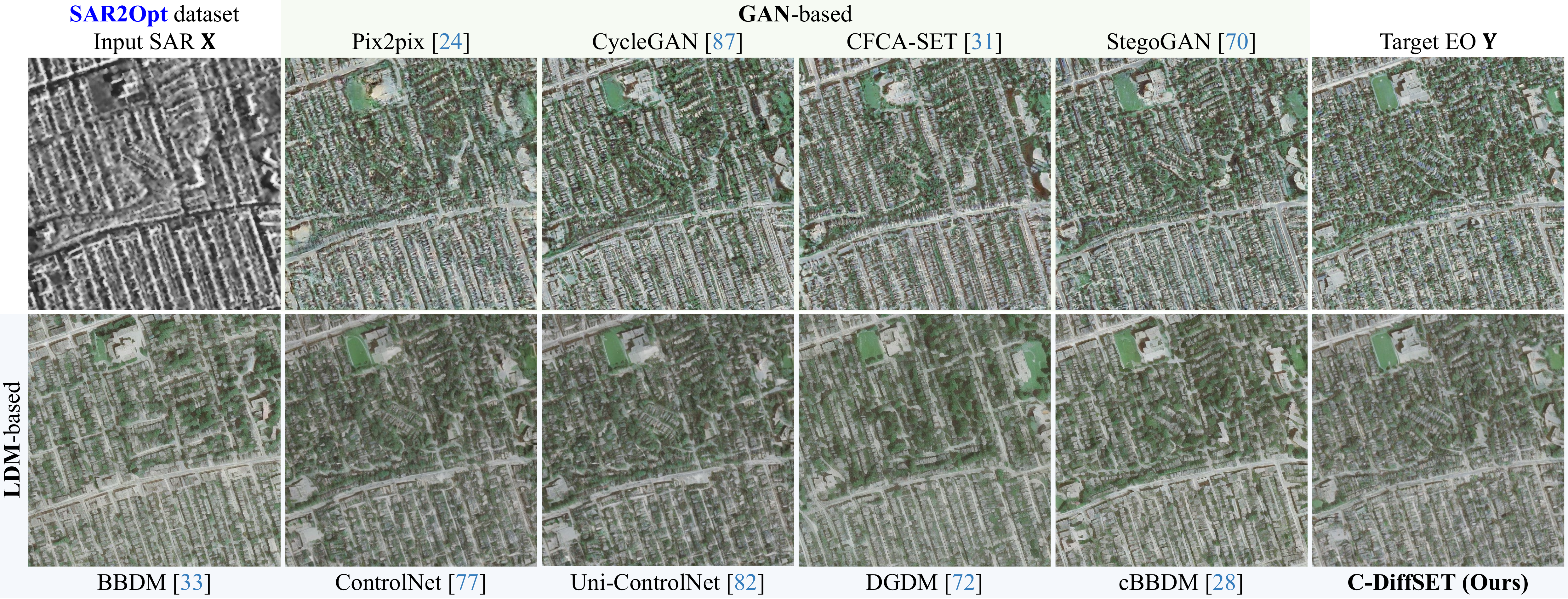}
  \caption{Visual comparison of SET results on the SAR2Opt dataset. 1st rows: GAN-based (Pix2pix, CycleGAN, CFCA-SET, and StegoGAN) methods. 2nd rows: LDM-based (BBDM, ControlNet, Uni-ControlNet, DGDM, cBBDM, and C-DiffSET) methods.}
  \label{fig:saropt_v1}
\end{figure*}

\begin{figure*}[tbp]
  \centering
  \includegraphics[width=1.0\textwidth]{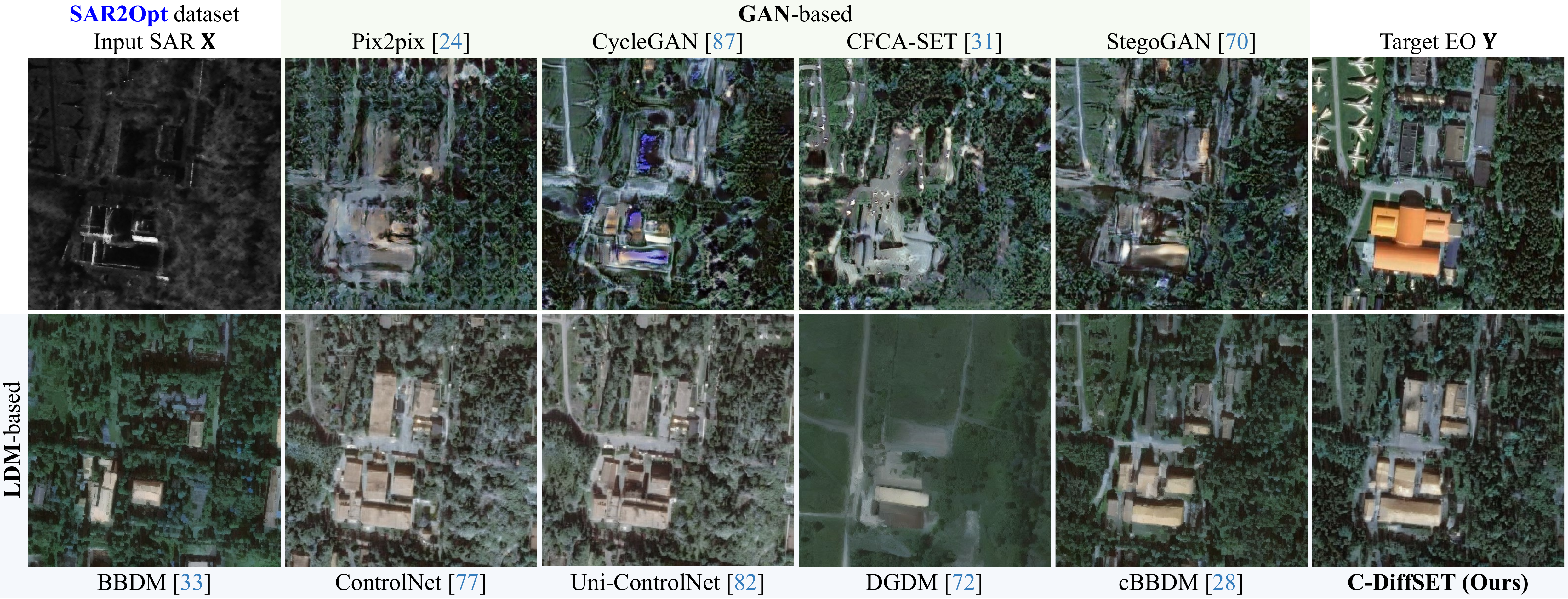}
  \caption{Visual comparison of SET results on the SAR2Opt dataset. 1st rows: GAN-based (Pix2pix, CycleGAN, CFCA-SET, and StegoGAN) methods. 2nd rows: LDM-based (BBDM, ControlNet, Uni-ControlNet, DGDM, cBBDM, and C-DiffSET) methods.}
  \label{fig:saropt_v2}
\end{figure*}

\begin{figure*}[tbp]
  \centering
  \includegraphics[width=1.0\textwidth]{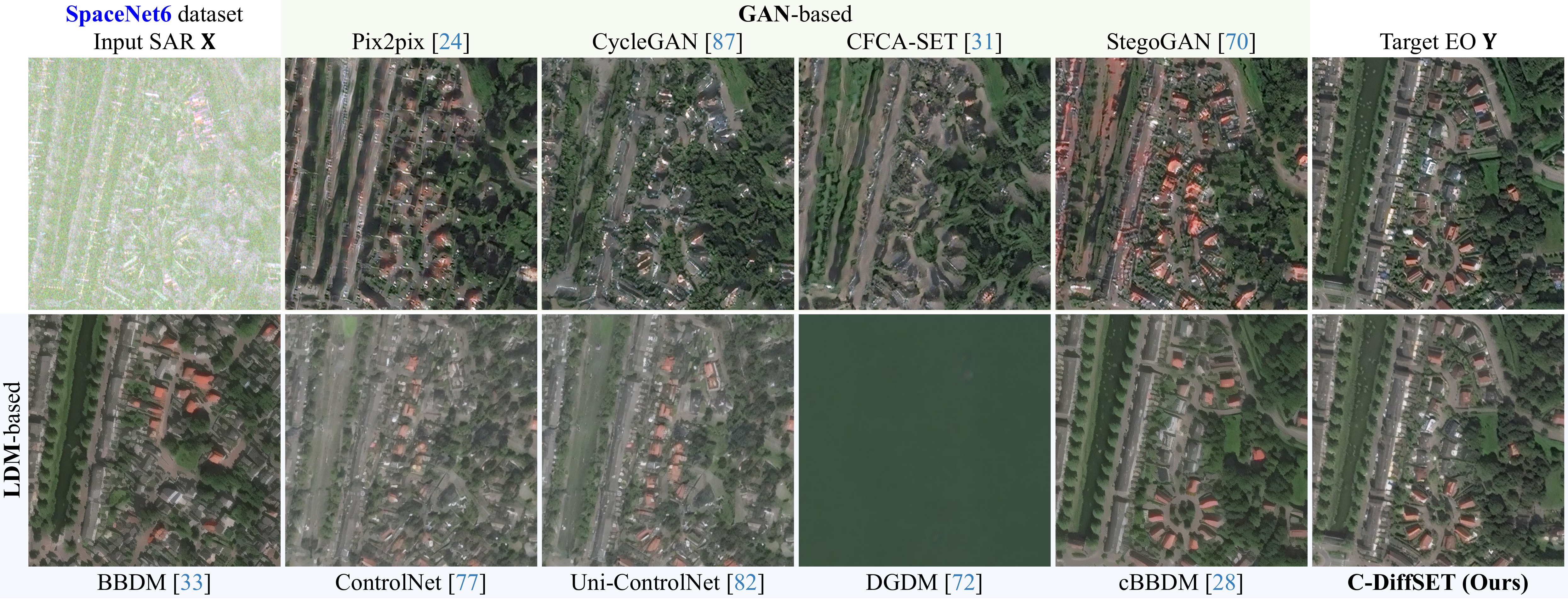}
  \caption{Visual comparison of SET results on the SpaceNet6 dataset. 1st rows: GAN-based (Pix2pix, CycleGAN, CFCA-SET, and StegoGAN) methods. 2nd rows: LDM-based (BBDM, ControlNet, Uni-ControlNet, DGDM, cBBDM, and C-DiffSET) methods.}
  \label{fig:spacenet_v1}
\end{figure*}

\begin{figure*}[tbp]
  \centering
  \includegraphics[width=1.0\textwidth]{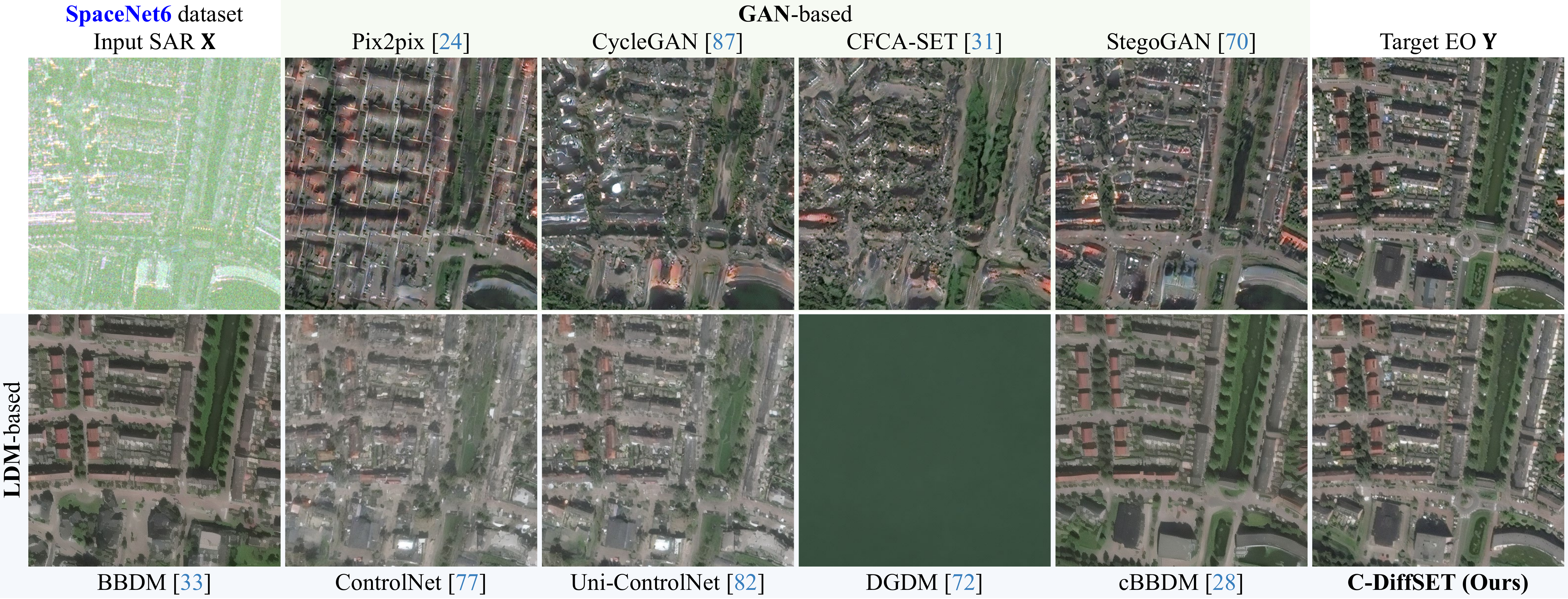}
  \caption{Visual comparison of SET results on the SpaceNet6 dataset. 1st rows: GAN-based (Pix2pix, CycleGAN, CFCA-SET, and StegoGAN) methods. 2nd rows: LDM-based (BBDM, ControlNet, Uni-ControlNet, DGDM, cBBDM, and C-DiffSET) methods.}
  \label{fig:spacenet_v2}
\end{figure*}

Fig.~\ref{fig:qxs_v5}, Fig.~\ref{fig:qxs_v1}, Fig.~\ref{fig:qxs_v2}, Fig.~\ref{fig:qxs_v3}, Fig.~\ref{fig:qxs_v4}, Fig.~\ref{fig:saropt_v1}, Fig.~\ref{fig:saropt_v2}, Fig.~\ref{fig:spacenet_v1}, and Fig.~\ref{fig:spacenet_v2} provide additional qualitative comparisons of SAR-to-EO image translation results on the QXS-SAROPT \cite{QXS}, SAR2Opt \cite{SAROpt}, and SpaceNet6 \cite{SpaceNet6} datasets. The GAN-based methods, including Pix2Pix \cite{pix2pix} and CycleGAN \cite{CycleGAN}, exhibit severe artifacts due to the inherent instability of the training process within the GAN frameworks. Although CFCA-SET \cite{CFCASET} and StegoGAN \cite{StegoGAN} mitigate some of these artifacts, they still produce visually inconsistent results, often failing to preserve fine-grained structural details. Among the LDM-based approaches, BBDM \cite{BBDM} and cBBDM \cite{cBBDM} generate smoother outputs and struggle with oversimplified textures and lack of structural alignment due to their direct diffusion process. DGDM \cite{DGDM} relies on an initial translator network to generate SAR-to-EO latents; however, the simplicity of this translator leads to poorly initialized latents, resulting in entirely unrealistic outputs. ControlNet-based approaches \cite{ControlNet, UniControlNet} exhibit similar limitations, as they keep the pretrained U-Net weights frozen and inject SAR conditions only in the decoder stage, failing to fully propagate the structural information of SAR throughout the denoising process. In contrast, our proposed C-DiffSET effectively addresses these limitations, producing visually coherent and structurally accurate EO images that are closely aligned with the target EO images.

\subsection{Analysis of Domain Gap and Upper Bound Performance}

In Fig.~\ref{fig:vae} of the main paper, we evaluate the reconstruction quality of the VAE encoder $\mathcal{E}_\text{vae}$ and decoder $\mathcal{D}_\text{vae}$ from the LDM by visualizing their outputs for target EO images and SAR images with varying levels of speckle noise.  

\begin{figure}[tbp]
  \centering
  \includegraphics[width=1.0\columnwidth]{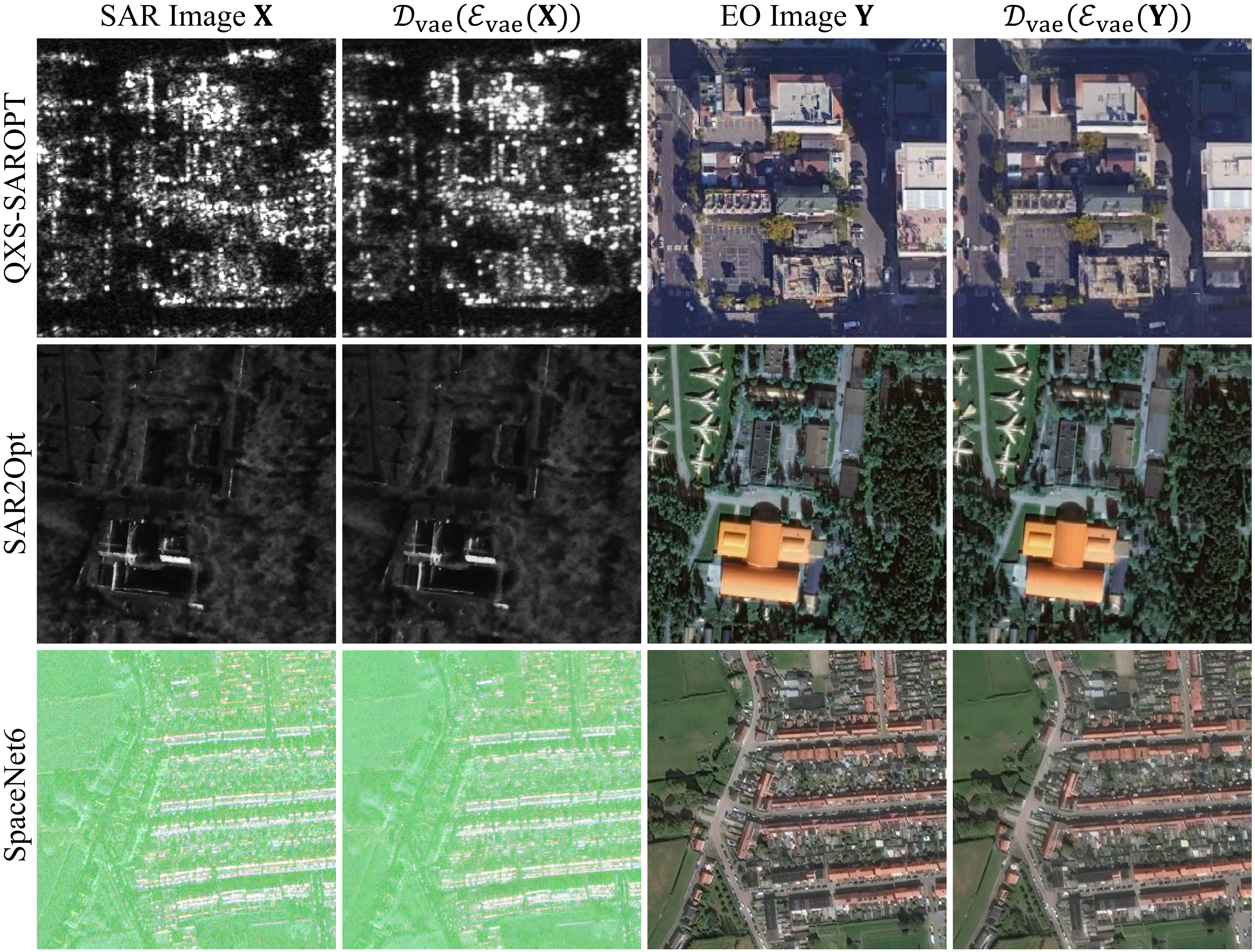}
  \caption{Results of applying the VAE encoder and decoder from LDM to EO and SAR images. These results suggest that the LDM VAE facilitates a shared latent representation between SAR and EO domains, supporting robust cross-domain image translation.}
  \label{fig:vae_supple}
  \vspace{-0.2cm}
\end{figure}

\noindent \textbf{Embedding into a shared latent space.}   
In Fig.~\ref{fig:vae_supple}, additional VAE reconstruction results demonstrate that SAR and EO images are embedded into the same latent space. Notably, VAE-reconstructed SAR images preserve structural information while reducing speckle noise, enhancing their suitability as conditioning inputs for SET task.

\noindent \textbf{Domain gap analysis.}
Fig.~\ref{fig:vae_supple}, Table~\ref{tab:bound}, Table~\ref{tab:bound2}, and Table~\ref{tab:bound3} show that the VAE, trained on natural image datasets, effectively reconstructs the RGB bands of EO images. This indicates that the domain gap between natural images and EO imagery in the RGB spectrum is relatively small, enabling the pretrained LDM to generalize well to EO data. This observation further supports our approach of leveraging a large-scale pretrained diffusion model for the SET task.

\begin{table}[tbp]
    \scriptsize
    \centering
    \resizebox{1.0\columnwidth}{!}{
    \def\arraystretch{1.2}
    \begin{tabular} {l|c|c|c|c|c}
        \Xhline{2\arrayrulewidth}
        \multirow{2}{*}{Methods} &\multicolumn{5}{c}{\textbf{SpaceNet6} Test Dataset} \\
        \cline{2-6}
        & FID$\downarrow$ & LPIPS$\downarrow$ & SCC$\uparrow$ & SSIM$\uparrow$ & PSNR$\uparrow$ \\
        \hline
        C-DiffSET (Ours) & 85.77 & 0.132 & 0.0210 & 0.546 & 21.498 \\
        $\mathcal{D}_\text{vae}(\mathcal{E}_\text{vae}(\mathbf{Y}))$ & 17.64 & 0.047 & 0.1417 & 0.800 & 28.543 \\
        \Xhline{2\arrayrulewidth}
    \end{tabular}}
    \caption{Comparison of C-DiffSET performance with the VAE upper bound on the SpaceNet6 dataset. The upper bound represents the maximum achievable quality defined by the VAE-decoded target EO features.}
    \label{tab:bound}
\end{table}

\begin{table}[tbp]
    \scriptsize
    \centering
    \resizebox{1.0\columnwidth}{!}{
    \def\arraystretch{1.2}
    \begin{tabular} {l|c|c|c|c|c}
        \Xhline{2\arrayrulewidth}
        \multirow{2}{*}{Methods} &\multicolumn{5}{c}{\textbf{SAR2Opt} Test Dataset} \\
        \cline{2-6}
        & FID$\downarrow$ & LPIPS$\downarrow$ & SCC$\uparrow$ & SSIM$\uparrow$ & PSNR$\uparrow$ \\
        \hline
        C-DiffSET (Ours) & 77.81 & 0.346 & 0.0035 & 0.286 & 16.613 \\
        $\mathcal{D}_\text{vae}(\mathcal{E}_\text{vae}(\mathbf{Y}))$ & 28.99 & 0.095 & 0.1583 & 0.667 & 25.678 \\
        \Xhline{2\arrayrulewidth}
    \end{tabular}}
    \caption{Comparison of C-DiffSET performance with the VAE upper bound on the SAR2Opt dataset.}
    \label{tab:bound2}
\end{table}

\begin{table}[tbp]
    \scriptsize
    \centering
    \resizebox{1.0\columnwidth}{!}{
    \def\arraystretch{1.2}
    \begin{tabular} {l|c|c|c|c|c}
        \Xhline{2\arrayrulewidth}
        \multirow{2}{*}{Methods} &\multicolumn{5}{c}{\textbf{QXS-SAROPT} Test Dataset} \\
        \cline{2-6}
        & FID$\downarrow$ & LPIPS$\downarrow$ & SCC$\uparrow$ & SSIM$\uparrow$ & PSNR$\uparrow$ \\
        \hline
        C-DiffSET (Ours) & 18.15 & 0.293 & 0.0108 & 0.372 & 18.077 \\
        $\mathcal{D}_\text{vae}(\mathcal{E}_\text{vae}(\mathbf{Y}))$ & 9.18 & 0.064 & 0.2597 & 0.794 & 29.497 \\
        \Xhline{2\arrayrulewidth}
    \end{tabular}}
    \caption{Comparison of C-DiffSET performance with the VAE upper bound on the QXS-SAROPT dataset.}
    \label{tab:bound3}
\end{table}

\begin{table}[tbp]
    \scriptsize
    \centering
    \resizebox{1.0\columnwidth}{!}{
    \def\arraystretch{1.2}
    \begin{tabular}{c|l|c|c|c|c}
        \Xhline{2\arrayrulewidth}
        Types & Methods & Params. (M) & FLOPs (G) & Memory (MB) & Time (s) \\
        \hline
        \multirow{5}{*}{GANs} & Pix2Pix \cite{pix2pix} & 54.41 & 24.22 & 464.12 & 0.06 \\
        & CycleGAN \cite{CycleGAN} & 7.84 & 140.43 & 398.38 & 0.08 \\
        & SAR-SMTNet \cite{youk2023transformer} & 2.15 & 615.40 & 2626.98 & 0.22 \\
        & CFCA-SET \cite{CFCASET} & 26.80 & 98.98 & 431.86 & 0.10 \\
        & StegoGAN \cite{StegoGAN} & 13.15 & 227.49 & 461.14 & 0.11 \\
        \hline
        \multirow{6}{*}{LDMs} & BBDM \cite{BBDM} & 949.56 & 2122.44 & 6147.88 & 3.13 \\
        & ControlNet \cite{ControlNet} & 1312.72 & 2231.03 & 7567.83 & 4.50 \\
        & Uni-ControlNet \cite{UniControlNet} & 1519.18 & 2295.20 & 8382.00 & 4.95 \\
        & DGDM \cite{DGDM} & 959.03 & 2161.25 & 6184.45 & 1.47 \\
        & cBBDM \cite{cBBDM} & 949.58 & 2122.49 & 6147.93 & 3.15 \\
        & \textbf{C-DiffSET} & 949.58 & 2122.49 & 6148.09 & 3.27 \\
        \Xhline{2\arrayrulewidth}
    \end{tabular}}
    \caption{Comparative analysis of C-DiffSET with other methods by parameters,
    FLOPs, memory usage, and inference time.}
    \label{tab:complex}
\end{table}

\begin{figure}[tbp]
  \centering
  \includegraphics[width=1.0\columnwidth]{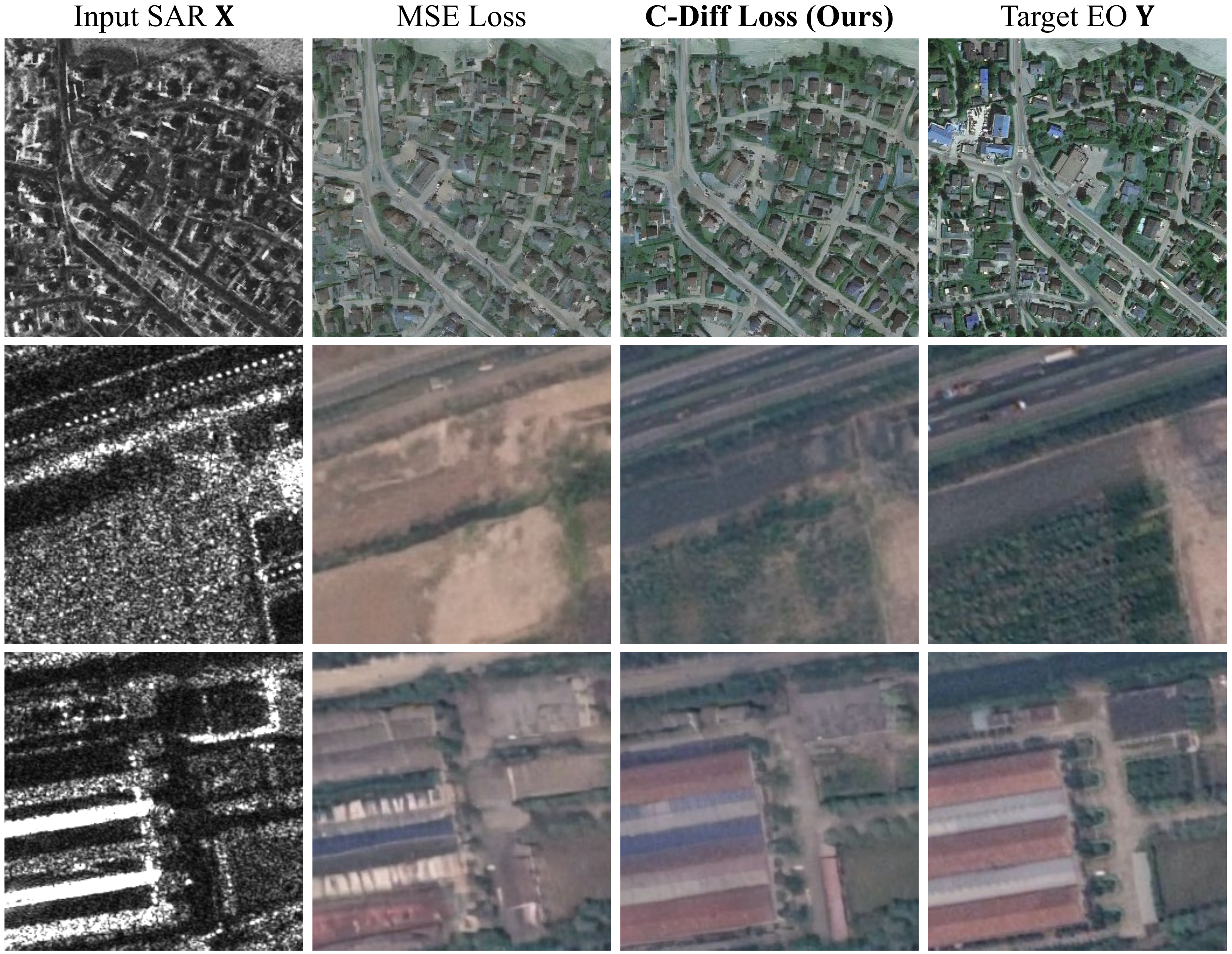}
  \caption{Visual comparison of SET results with and without C-Diff loss. The MSE loss corresponds to $\beta=0$, indicating no confidence weighting.}
  \label{fig:ablation_loss}
\end{figure}

\begin{figure*}[tbp]
  \centering
  \includegraphics[width=1.0\textwidth]{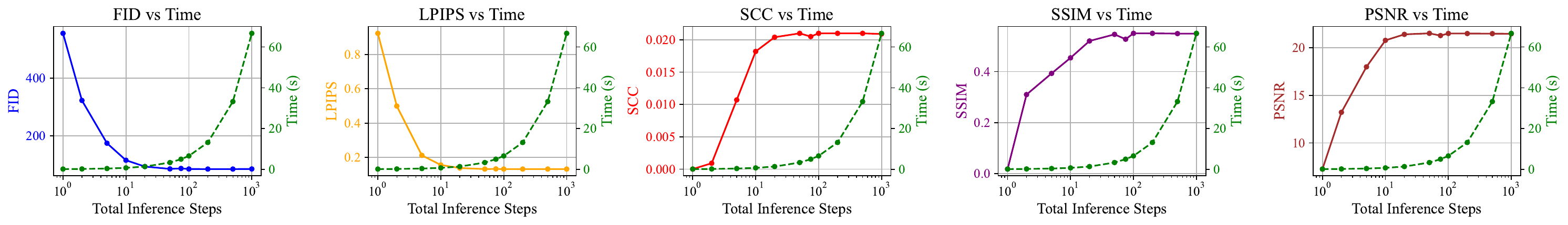}
  \caption{Impact of total inference steps on performance metrics (FID, LPIPS, SCC, SSIM, and PSNR) and inference time.}
  \label{fig:graph}
\end{figure*}

\begin{figure*}[tbp]
  \centering
  \includegraphics[width=1.0\textwidth]{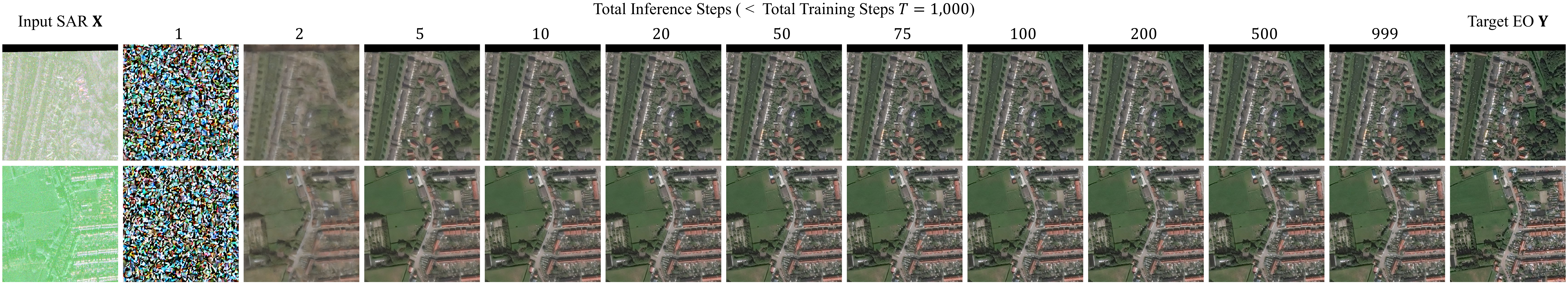}
  \caption{Visualization of C-DiffSET results across varying numbers of total inference steps.}
  \label{fig:inference}
\end{figure*}

\begin{figure*}[tbp]
  \centering
  \includegraphics[width=1.0\textwidth]{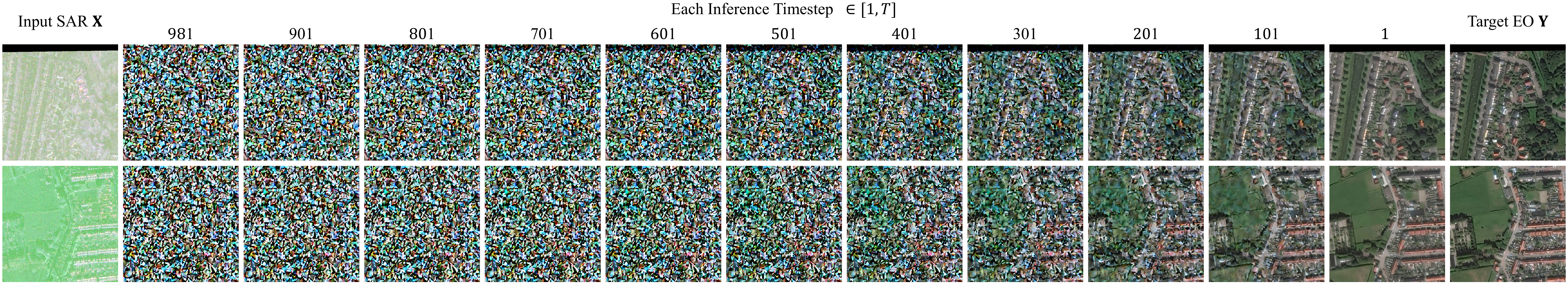}
  \caption{Visualization of C-DiffSET results across inference timesteps on the SpaceNet6 dataset with a total of 50 inference steps. Each column corresponds to an inference step, starting with substantial noise at $t=981$ and progressively refining the output to match the target EO image at $t=1$.}
  \label{fig:denoise}
\end{figure*}

\begin{table}[tbp]
    \scriptsize
    \centering
    \resizebox{1.0\columnwidth}{!}{
    \def\arraystretch{1.2}
    \begin{tabular} {c|c|c|c|c|c|c}
        \Xhline{2\arrayrulewidth}
        \multirow{2}{*}{\makecell{Pretrained\\LDM}} & \multirow{2}{*}{\makecell{Loss\\function}} &\multicolumn{5}{c}{ \textbf{QXS-SAROPT} Dataset} \\
        \cline{3-7}
        & & FID$\downarrow$ & LPIPS$\downarrow$ & SCC$\uparrow$ & SSIM$\uparrow$ & PSNR$\uparrow$ \\
        \hline
        & MSE & 29.04 & 0.407 & 0.0006 & 0.279 & 14.647 \\
        \checkmark & MSE & 19.99 & 0.297 & 0.0094 & 0.364 & 17.736 \\
        \checkmark & C-Diff & {\color{red}{\textbf{18.15}}} & {\color{red}{\textbf{0.293}}} & {\color{red}{\textbf{0.0108}}} & {\color{red}{\textbf{0.372}}} & {\color{red}{\textbf{18.077}}} \\
        \Xhline{2\arrayrulewidth}
    \end{tabular}}
    \caption{Ablation studies on the QXS-SAROPT dataset evaluating the impact of pretrained LDM and confidence-guided diffusion (C-Diff) loss.}
    \label{tab:more_ablation}
\end{table}

\begin{table}[tbp]
    \scriptsize
    \centering
    \resizebox{1.0\columnwidth}{!}{
    \def\arraystretch{1.2}
    \begin{tabular} {c|c|c|c|c|c}
        \Xhline{2\arrayrulewidth}
        \multirow{2}{*}{Text Prompt} &\multicolumn{5}{c}{\textbf{SpaceNet6} Dataset} \\
        \cline{2-6}
        & FID$\downarrow$ & LPIPS$\downarrow$ & SCC$\uparrow$ & SSIM$\uparrow$ & PSNR$\uparrow$ \\
        \hline
        ``\;\;" (Null text, $\varnothing$) & 79.01 & 0.351 & 0.0027 & 0.273 & 16.237 \\
        ``Eletro-Optical Image"& {\color{red}{\textbf{77.81}}} & {\color{red}{\textbf{0.346}}} & {\color{red}{\textbf{0.0035}}} & {\color{red}{\textbf{0.286}}} & {\color{red}{\textbf{16.613}}} \\
        \Xhline{2\arrayrulewidth}
    \end{tabular}}
    \caption{Ablation studies on the text prompts.}
    \label{tab:text}
\end{table}

\noindent \textbf{Upper bound comparison.}
Our proposed C-DiffSET framework, built on the LDM architecture, predicts the target EO features $\mathbf{z}_y = \mathcal{E}_\text{vae}(\mathbf{Y})$ embedded in the latent space through the VAE encoder. The VAE-decoded reconstruction of the target EO feature, $\mathcal{D}_\text{vae}(\mathbf{z}_y)=\mathcal{D}_\text{vae}(\mathcal{E}_\text{vae}(\mathbf{Y}))$, serves as an upper bound for our C-DiffSET, representing the maximum achievable quality given the pretrained LDM. Table~\ref{tab:bound}, Table~\ref{tab:bound2}, and Table~\ref{tab:bound3} quantitatively compare the performance of C-DiffSET against this upper bound. Although our framework achieves strong results, particularly in perceptual metrics such as LPIPS and FID, the gap to the upper bound highlights the inherent challenges in SAR-to-EO translation, including noise, misalignment, and the structural complexity of SAR data. This analysis underscores the potential for future research to bridge this gap, pushing SET closer to the theoretical upper limit.

\subsection{Computational Complexity}

We compare the number of parameters, FLOPs, memory usage, and inference time for 512$\times$512 images in Table~\ref{tab:complex}. As noted, the LDM-based methods are more complex than the GAN-based ones. The time-performance trade-off is in Fig.~\ref{fig:graph}.

\subsection{Progressive Denoising Visualization}

Fig.~\ref{fig:denoise} illustrates the progression of generated images across different inference timesteps during the reverse denoising process. Out of the total $T_\text{test}=50$ inference steps, we select 11 representative timesteps to visualize the progressive refinement of the output. Using the DDIM \cite{DDIM} noise scheduler, denoising is performed over 50 steps, with timesteps chosen from the range $[1, T]$ to balance sampling efficiency and performance. At earlier timesteps (e.g., $t=981$), the images exhibit significant noise, reflecting the initial latent representation. As the process progresses to later timesteps (e.g., $t=1$), the generated images become increasingly coherent, closely resembling the target EO images.

\section{Further Ablation Studies}

\subsection{Impact of Pretrained LDM and C-Diff Loss}

We provide in Table~\ref{tab:more_ablation} further experiments on additional dataset to validate $\mathcal{L}_\text{C-Diff}$. Moreover, Fig.~\ref{fig:ablation_loss} presents visual comparisons of SET results under different loss functions, demonstrating the impact of $\mathcal{L}_\text{C-Diff}$ on structural consistency and perceptual quality.

\subsection{Effect of Text Prompts}

We provide additional ablation studies on the text prompt in Table~\ref{tab:text}. Since our datasets do not include text annotations, we use a generic prompt (fixed for all samples), with Stable Diffusion v2.1’s classifier-free guidance rather than a null prompt $\varnothing$.

\subsection{Effect of Total Inference Steps}

In Fig.~\ref{fig:graph} and Fig.~\ref{fig:inference}, we justify the selection of $T_\text{test}=50$ inference steps in our experiments. While setting the total inference steps to match the training steps ($T = 1,000$) ensures high performance, it incurs prohibitive inference times. To balance performance and efficiency, we evaluate the tradeoff between inference time and metric performance by varying the DDIM noise scheduler's inference steps. For computational feasibility, experiments are conducted on the first 50 samples of the SpaceNet6 dataset. The results indicate significant performance gains for lower step counts ($T_\text{test} = 1$ to $T_\text{test} = 50$), but diminishing returns beyond 50 steps, with negligible improvements in metrics such as FID, LPIPS, and PSNR. Based on this analysis, we select a total of 50 inference steps as the optimal trade-off, achieving high-quality outputs within practical inference times (3.376 seconds per 512$\times$512 image).

\section{Additional Experimental Details}

Since SAR features were incorporated as additional inputs to the U-Net $\psi$, the weights of the $\psi$’s first convolutional layer were initialized by repeating the original weights across the SAR channels. Additionally, the confidence map was initialized to 1 to ensure stable optimization, starting from the standard $\ell_2$ loss, and $\tau = \log2\pi$ \cite{seitzer2022pitfalls} was applied. For the QXS-SAROPT dataset, we used the provided original 256$\times$256-sized patches without modification, with a batch size of 64. For the SAR2Opt and SpaceNet6 datasets, we performed random cropping to 512$\times$512-sized patches and set the batch size to 16. Data augmentation techniques, including random horizontal and vertical flips and random rotations (multiples of 90 degrees), were applied during training to improve generalization. To prevent overfitting and mitigate pretrained weight forgetting, we used a small initial learning rate of $3 \times 10^{-5}$. To ensure reproducibility, a random seed of 2,025 was fixed for all experiments. Data was split with 80\% used for training and 20\% for testing across all experiments. 

\clearpage
{
    \small
    \bibliographystyle{ieeenat_fullname}
    \bibliography{main}
}
\end{document}